\documentclass[english]{article}
\usepackage[T1]{fontenc}
\usepackage[latin9]{inputenc}
\usepackage{babel}
\usepackage{amsmath}
\usepackage{amssymb}
\usepackage{graphicx}
\usepackage{esint}
\usepackage[unicode=true]
 {hyperref}

\makeatletter
\PassOptionsToPackage{bookmarks=false}{hyperref}

\@ifundefined{showcaptionsetup}{}{%
 \PassOptionsToPackage{caption=false}{subfig}}
\usepackage{subfig}
\makeatother

\begin{document}

\title{Optimal Encoding and Decoding for Point Process Observations: an
Approximate Closed-Form Filter}

\author{Yuval Harel\\
Department of Electrical Engineering\\
Technion -- Israel Institute of Technology\\
Technion City, Haifa, Israel\\
\href{mailto:yharel@tx.technion.ac.il}{yharel@tx.technion.ac.il}\\
\and Ron Meir\\
Department of Electrical Engineering\\
Technion -- Israel Institute of Technology\\
Technion City, Haifa, Israel\\
\href{mailto:rmeir@ee.technion.ac.il}{rmeir@ee.technion.ac.il}\\
\and Manfred Opper\\
Department of Electrical Engineering and Computer Science\\
Technical University Berlin\\
Berlin 10587, Germany\\
\href{mailto:manfred.opper@tu-berlin.de}{manfred.opper@tu-berlin.de}}
\maketitle
\begin{abstract}
The process of dynamic state estimation (filtering) based on point
process observations is in general intractable. Numerical sampling
techniques are often practically useful, but lead to limited conceptual
insight about optimal encoding/decoding strategies, which are of significant
relevance to Computational Neuroscience. We develop an analytically
tractable Bayesian approximation to optimal filtering based on point
process observations, which allows us to introduce distributional
assumptions about sensor properties, that greatly facilitate the analysis
of optimal encoding in situations deviating from common assumptions
of uniform coding. Numerical comparison with particle filtering demonstrate
the quality of the approximation. The analytic framework leads to
insights which are difficult to obtain from numerical algorithms,
and is consistent with biological observations about the distribution
of sensory cells' tuning curve centers.
\end{abstract}
\global\long\def\E{\mathrm{E}}
\global\long\def\mc#1{\mathcal{#1}}

\section{Introduction}

A key task facing natural or artificial agents acting in the real
world is that of causally inferring a hidden dynamic state based on
partial noisy observations. This task, referred to as \emph{filtering}
in the engineering literature, has been extensively studied since
the 1960s, both theoretically and practically (e.g., \cite{AndMoo05}).
For the linear setting with Gaussian noise and quadratic cost, the
solution is well known both for discrete and continuous times, leading
to the celebrated Kalman and the Kalman-Bucy filters \cite{KalBuc61,Kalman60},
respectively. In these cases the exact posterior state distribution
is Gaussian, resulting in closed form recursive update equations for
the mean and variance, yielding finite-dimensional filters. However,
beyond some very specific settings \cite{Daum05}, the optimal filter
is infinite-dimensional and impossible to compute in closed form,
requiring either approximate analytic techniques (e.g., the extended
Kalman filter (e.g., \cite{AndMoo05}), the unscented filter \cite{JulUhl00},
the cubature filter \cite{Arasaratnam2009}) or numerical procedures
(e.g., particle filters \cite{DouJoh09}). The latter usually require
time discretization and a finite number of particles, resulting in
loss of precision in a continuous time context. 

The present work is motivated by the increasingly available data from
neuroscience, where the spiking activity of sensory neurons gives
rise to Point Process (PP)-like activity, which must be further analyzed
and manipulated by networks of neurons, in order to estimate attributes
of the external environment (e.g., the location and velocity of an
object), or to control the body, as in motor control, based on visual
and proprioceptive sensory inputs. In both these cases the system
faces the difficulty of assessing, in real-time, the environmental
state through a large number of simple, restricted and noisy sensors
(e.g., \cite{DayAbb05}). Example of such sensory cells are retinal
ganglion cells, auditory (cochlear) cells and proprioceptive stretch
receptors. In all these cases, information is conveyed to higher brain
areas through sequences of sharp pulses (spikes) delivered by the
responses of large numbers of sensory cells (about a million such
cells in the visual case). Each sensory cell is usually responsive
to a narrow set of attributes of the external stimuli (e.g., positions
in space, colors, frequencies etc.). Such cells are characterized
by \emph{tuning functions} or \emph{tuning curves}, with differential
responses to attributes of the external stimulus. For example, a visual
cell could respond with maximal probability to a stimulus at a specific
location in space and with reduced probability to stimuli distanced
from this point. An auditory cell could respond strongly to a certain
frequency range and with diminished responses to other frequencies,
etc. In all these cases, the actual firing of cells is random (due
to stochastic elements in the neurons, e.g., ion channels and synapses),
and can be described by PP with rate determined by the input and by
the cell's tuning function \cite{DayAbb05}. A particularly important
and ubiquitous phenomenon in biological sensory systems is \emph{sensory
adaptation}, which is the stimulus-dependent modification of system
parameters in a direction enhancing performance. Such changes usually
take place through the modification of tuning function properties,
e.g., the narrowing of tuning curve widths \cite{Benucci2009,GutZheKnu02},
the change of tuning curve heights \cite{Benucci2013}, or the shift
of the center position of tuning curves \cite{HarMcAlp04}. In the
sequel we refer to the process of setting the neurons' sensory tuning
functions as \emph{encoding}, and to the process of reconstructing
the state based on the PP observations as \emph{decoding}. 

Inferring the hidden state under such circumstances has been widely
studied within the Computational Neuroscience literature, mostly for
static stimuli, homogeneous and equally-spaced tuning functions, and
using various approximations to the reconstruction error, such as
the Fisher information. In this work we are interested in setting
up a general framework for PP filtering in continuous time, and establishing
closed-form analytic expressions for an approximately optimal filter
(see \cite{BobMeiEld09,SusMeiOpp11,SusMeiOpp13} for previous work
in related, but more restrictive, settings). We aim to characterize
the nature of near-optimal encoders, namely to determine the structure
of the tuning functions for optimal state inference. A significant
advantage of the closed form expressions over purely numerical techniques
is the insight and intuition that is gained from them about \textit{qualitative}
aspects of the system. Moreover, the leverage gained by the analytic
computation contributes to reducing the variance inherent to Monte
Carlo approaches. Note that while this work has been motivated by
neuroscience, it should be viewed as a contribution to the theory
of point process filtering. 

Technically, given the intractable infinite-dimensional nature of
the posterior distribution, we use a projection method replacing the
full posterior at each point in time by a projection onto a simple
family of distributions (Gaussian in our case). This approach, originally
developed in the Filtering literature \cite{Maybeck79,BriHanLeg98},
and termed Assumed Density Filtering (ADF), has been successfully
used more recently in Machine Learning \cite{Opper98,Minka01}. We
are aware of a single previous work using ADF in the context of point
process observations (\cite{SusemihlThesis2014}), where it was used
to optimize encoding by a population of two neurons. 

Filtering PP observations based on multi-variate dynamic states in
continuous time has received far less attention in the engineering
literature than filtering based on more standard observations. While
a stochastic PDE for the infinite-dimensional posterior state distribution
can be derived under general conditions \cite{Snyder1972} (see also
\cite{Segall1976}), these equations are intractable in general, and
cannot even be qualitatively analyzed. This led \cite{RhoSny1977}
to consider a special case where the rate functions for the PP were
homogeneous Gaussians, leading to a posterior Gaussian distribution
of the state, giving rise to simple exact filtering equations for
the posterior mean and variance of the finite-dimensional posterior
state distribution. However, in our present setting we are motivated
to study inhomogeneous rate functions which lead to non-Gaussian infinite-dimensional
posterior distributions necessitating the introduction of finite-dimensional
approximations. Beyond neuroscience, PP based filtering has been used
for position sensing and tracking in optical communication \cite[sec. 4]{Snyder1977},
control of computer communication networks \cite{Segall1978}, queuing
\cite{Bremaud81} and econometrics \cite{Frey2001}, although our
main motivation for studying optimal sensory encoding arises from
neuroscience. Based on this motivation, we sometimes refer to point
events as ``spikes''.

The main contributions of the paper are the following: \emph{(i)}
Derivation of closed form recursive expressions for the continuous
time posterior mean and variance within the ADF approximation, allowing
for the incorporation of distributional assumptions over sensory variables
(going beyond homogeneity assumptions used so far). \emph{(ii)} Demonstrating
the quality of the ADF approximation by comparison to state of the
art particle filtering simulation methods. \emph{(iii)} Characterization
of optimal adaptation (encoding) for sensory cells in a more general
setting than hitherto considered. \emph{(iv)} Demonstrating the interesting
interplay between prior information and neuronal firing, showing how
in certain situations, the absence of spikes can be highly informative.
Preliminary results discussed in this paper were presented at a conference
\cite{Harel2015}. The present paper provides a rigorous formulation
of the mathematical framework and closed-form expressions for cases
that were not discussed in \cite{Harel2015}, including finite, possibly
heterogeneous, mixtures of Gaussian and uniform population.

\section{Problem Formulation}

\subsection{Heuristic formulation}

\begin{figure}
\bigskip{}

\includegraphics[width=1\columnwidth]{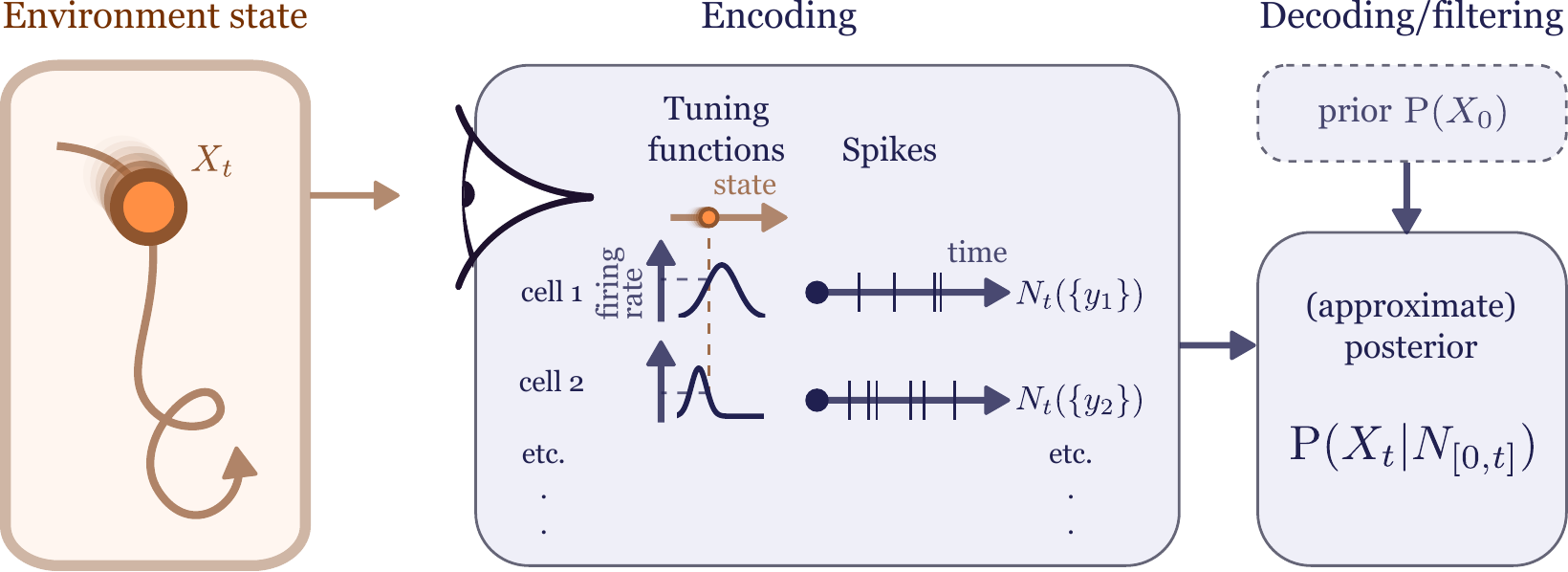}

\bigskip{}
\caption{Problem setting}
\label{setting}
\end{figure}

We consider a dynamical system with state $X_{t}\in\mathbb{R}^{n}$,
for $t\in[0,\infty)$, observed through an array of sensors. Heuristically,
we assume the $i$-th sensor generates a random point in response
to $X_{t}=x$ with probability $\lambda_{t}\left(x;y_{i}\right)dt$
in the time interval $[t,t+dt)$, where $y_{i}\in\mathbf{Y}$ is a
parameter of the sensor. For example, in the case of auditory neurons,
$y_{i}$ may be the tuning curve center of the $i$-th neuron, corresponding
to the frequency for which it responds with the highest firing rate.
In this case the space $\mathbf{Y}$ would be the space of frequencies. 

We assume that given $X_{t}=x$, points are generated independently
of the past of $X$ and of previously generated points from all sensors.
Each generated point is ``marked'' with the parameter $y_{i}$ of
the sensor that generated it. We therefore describe the observations
by a marked point process, i.e., a random counting measure $N$ on
$[0,\infty)\times\mathcal{\mathbf{Y}}$ where $\mathbf{Y}$ is the
\emph{mark space}, and the observations available at time $t$ are
the restriction of $N$ to the set $[0,t]\times\mathcal{\mathbf{Y}}$.
Such a process may also be described as a vector of (unmarked) point
processes, one for each sensor; however, we adopt the marked point
process view to allow taking the limit of infinitely many sensors,
as described below. Figure \ref{setting} illustrates this setting
in a biological context, where the sensors are neurons and the points
events are spikes (action potentials).

We denote by $f$ the counting measure of sensor marks over the mark
space $\mathbf{Y}$, i.e., $f=\sum_{i}\delta_{y_{i}}$, where $\delta_{y}$
is the Dirac delta measure at $y$. Then the marked process $N_{t}$
has the measure-valued random process $\lambda_{t}\left(X_{t};y\right)f\left(dy\right)$
as its \emph{intensity kernel} (see, e.g., \cite{Bremaud81}, Chapter
VIII), meaning that the rate (or \emph{intensity}) of points with
marks in a set $Y\subseteq\mathbf{Y}$ at time $t$ is
\[
\sum_{i:y_{i}\in Y}\lambda_{t}\left(X_{t};y_{i}\right)=\int_{Y}\lambda_{t}\left(X_{t};y\right)f\left(dy\right).
\]
Thus, the dynamics of $N$ may be described heuristically by means
of the intensity kernel as 
\begin{equation}
\E\left[N\left(dt,dy\right)|X_{\left[0,t\right]},N_{\left[0,t\right]}\right]=\lambda_{t}\left(X_{t};y\right)f\left(dy\right)dt.\label{eq:rate-heuristic}
\end{equation}
Continuing the auditory example, the expected instantaneous total
rate of spikes from all neurons which are tuned to a frequency within
the range $\left[y_{\mathrm{min}},y_{\mathrm{max}}\right]$ (conditioned
on the history of the external state and of neural firing) would be
given by the integral of $\lambda_{t}\left(X_{t};y\right)f\left(dy\right)$
over this range of frequencies.

We generalize this model by allowing $f$ to be an arbitrary measure
on $\mathbf{Y}$ (not necessarily discrete). A continuous measure
may be useful as an approximate model for the response of a large
number of sensors, which may be easier to analyze than its discrete
counterpart. For example, a simplifying assumption in some previous
works is that tuning curve centers are located on an equally-spaced
grid (e.g., \cite{RhoSny1977,YaeMei10,Susemihl2014}). In the limit
of infinitely many neurons, this is equivalent to taking $f$ to be
the Lebesgue measure\textemdash the uniform ``distribution'' over
the entire space.

\subsection{Rigorous formulation}

We now proceed to describe our model more rigorously. We assume the
state $X_{t}\in\mathbb{R}^{n}$ obeys a Stochastic Differential Equation
(SDE)
\begin{equation}
dX_{t}=A\left(X_{t}\right)dt+B\left(U_{t}\right)dt+D\left(X_{t}\right)dW_{t},\quad\left(t\geq0\right),\label{eq:dynamics}
\end{equation}
where $A\left(\cdot\right),B\left(\cdot\right),D\left(\cdot\right)$
are arbitrary functions such that the SDE has a unique solution, $U$
is a control process and $W$ is a Wiener process. The initial condition
$X_{0}$ is assumed to have a continuous distribution with a known
density.

The observation process $N$ is a marked point process with marks
from a measurable space $\left(\mathbf{Y},\mathcal{Y}\right)$, i.e.,
a random counting measure on $[0,\infty)\times\mathbf{Y}$. We use
the notation $N_{t}\left(Y\right)\triangleq N\left([0,t]\times Y\right)$
for $Y\in\mathcal{Y}$, and when $Y=\mathbf{Y}$ we omit it and write
simply $N_{t}$.

Denote by $\left(\Omega,\mathcal{F},P\right)$ the underlying probability
space, and by $\mathcal{X},\mathcal{N}$ the filtrations generated
by $X,N$ respectively, that is, 
\begin{align*}
\mathcal{X}_{t} & \triangleq\sigma\left(X_{\left[0,t\right]}\right)=\sigma\left\{ X_{s}:s\leq t\right\} \\
\mathcal{N}_{t} & \triangleq\sigma\left(N_{\left[0,t\right]\times\mathbf{Y}}\right)=\sigma\left\{ N\left((a,b]\times Y\right):a<b\leq t,Y\in\mathcal{Y}\right\} 
\end{align*}
 where $N_{\left[0,t\right]\times\mathbf{Y}}$ is the measure $N$
restricted to $\left[0,t\right]\times\mathbf{Y}$. We assume the control
process $U$ is adapted to $\mathcal{N}$. Let $\mathcal{B}_{t}=\mathcal{X}_{t}\vee\mathcal{N}_{t}$
be the smallest $\sigma$-algebra containing $\mathcal{X}_{t}$ and
$\mathcal{N}_{t}$. We assume that the process $N_{t}\left(Y\right)$
has an intensity kernel with respect to $\mathcal{B}$ given by $\lambda_{t}\left(X_{t};y\right)f\left(dy\right)$
for some known function $\lambda_{t}\left(x;y\right)$ and measure
$f\left(dy\right)$ on $\mathcal{Y}$, meaning that for each $Y\in\mathcal{Y}$,
$C_{t}^{Y}\triangleq\int_{0}^{t}\left(\int_{Y}\lambda_{s}\left(X_{s};y\right)f\left(dy\right)\right)ds$
is $\mathcal{B}_{t}$-predictable and $N_{t}\left(Y\right)-C_{t}^{Y}$
is a $\mathcal{B}_{t}$-martingale. This condition is the rigorous
equivalent of (\ref{eq:rate-heuristic}) (see, e.g., \cite{Segall1975-modelling}
for a discussion of this definition in the unmarked case). Note in
particular that the presence of feedback in (\ref{eq:dynamics}),
which may depend on the history of $N$, means that $N\left(Y\right)$
is not in general a doubly-stochastic Poisson process, which forces
us to resort to a more sophisticated definition. Our results are applicable
to this general setting, and not restricted in application to doubly-stochastic
Poisson processes.

We define 
\begin{align*}
\hat{\lambda}_{t}\left(y\right) & \triangleq\E\left[\lambda_{t}\left(X_{t};y\right)|\mathcal{N}_{t}\right],\\
\hat{\lambda}_{t}^{f} & \triangleq\int\hat{\lambda}_{t}\left(y\right)f\left(dy\right).
\end{align*}
The measure $\hat{\lambda}_{t}\left(y\right)f\left(dy\right)$ is
the intensity kernel of $N$ with respect to its natural filtration
$\mathcal{N}$, i.e., $\int_{Y}\hat{\lambda}_{t}\left(y\right)f\left(dy\right)$
is the intensity of $N_{t}\left(Y\right)$ w.r.t. $\mathcal{N}$ for
each $Y\in\mathcal{Y}$ (see \cite{Segall1975-modelling}, Theorem
2), and $\hat{\lambda}_{t}^{f}$ is the intensity of the unmarked
process $N_{t}$. For $Y\in\mathcal{Y}$, the innovation process of
$N_{t}\left(Y\right)$ is therefore $N_{t}\left(Y\right)-\int_{0}^{t}\int_{Y}\hat{\lambda}_{t}\left(y\right)f\left(dy\right)ds$,
and accordingly we define the \emph{innovation measure} $I$ by
\begin{equation}
I\left(dt,dy\right)\triangleq N\left(dt,dy\right)-\hat{\lambda}_{t}\left(y\right)f\left(dy\right)dt.\label{eq:innovation}
\end{equation}

\subsection{Encoding and decoding\label{sub:Encoding-and-decoding}}

We consider the question of optimal encoding and decoding under the
above model. By \emph{decoding }we mean computing (exactly or approximately)
the full posterior distribution of $X_{t}$ given $\mathcal{N}_{t}$.
The problem of optimal encoding is then the problem of optimal sensor
configuration, i.e., finding the optimal rate function $\lambda_{t}\left(x;y\right)$
and population distribution $f\left(dy\right)$ so as to minimize
some performance criterion. We assume the $\lambda_{t}\left(x;y\right)$
and $f\left(dy\right)$ come from some parameterized family with parameter
$\phi$.

To quantify the performance of the encoding-decoding system, we summarize
the result of decoding using a single estimator $\hat{X}_{t}=\hat{X}_{t}\left(\mathcal{N}_{t}\right)$,
and define the Mean Square Error (MSE) as $\epsilon_{t}\triangleq\mathrm{trace}[(X_{t}-\hat{X}_{t})(X_{t}-\hat{X}_{t})^{T}]$.
We seek $\hat{X}_{t}$ and $\phi$ that solve $\min_{\phi}\min_{\hat{X}_{t}}\mathrm{E}\left[\epsilon_{t}\right]=\min_{\phi}\mathrm{E}[\min_{\hat{X}_{t}}\mathrm{E}[\epsilon_{t}|\mathcal{N}_{t}]]$.
The inner minimization problem in this equation is solved by the MSE-optimal
decoder, which is the posterior mean $\hat{X}_{t}=\mu_{t}\triangleq\mathrm{E}\left[X_{t}|\mathcal{N}_{t}\right]$.
The posterior mean may be computed from the full posterior obtained
by decoding. The outer minimization problem is solved by the optimal
encoder. Note that, although we assume a fixed parameter $\phi$ which
does not depend on time, the optimal value of $\phi$ for which the
minimum is obtained generally depends on the time $t$ where the error
is to be minimized. In principle, the encoding/decoding problem can
be solved for any value of $t$. In order to assess performance it
is convenient to consider the steady-state limit $t\to\infty$ for
the encoding problem.

Below, we approximately solve the decoding problem for any $t$, for
some specific forms of $\lambda_{t}$ and $f$. We then explore the
problem of choosing the steady-state optimal encoding parameters $\phi$
using Monte Carlo simulations in a biologically-motivated toy model.
Note that if decoding is exact, the problem of optimal encoding becomes
that of minimizing the expected posterior variance.

Having an efficient (closed-form) approximate filter allows performing
the Monte Carlo simulation at a significantly reduced computational
cost, relative to numerical methods such as particle filtering. The
computational cost is further reduced by averaging the computed posterior
variance across trials, rather than the squared error, thereby requiring
fewer trials. The mean of the posterior variance equals the MSE (of
the posterior mean), but has the advantage of being less noisy than
the squared error itself \textendash{} since by definition it is the
mean of the square error under conditioning on $\mathcal{N}_{t}$
.

\subsection{Special cases\label{sub:special-cases}}

To approximately solve the decoding problem in closed form, we focus
on the case of Gaussian sensors: each sensor is marked with $y=\left(h,\theta,R\right)$
where $h\in\mathbb{R}_{+},\theta\in\mathbb{R}^{m}$, $R\in\mathbb{R}^{m\times m}$
is positive semidefinite, $m\leq n$, and 
\begin{equation}
\lambda_{t}\left(x;h,\theta,R\right)=h\exp\left(-\frac{1}{2}\left\Vert Hx-\theta\right\Vert _{R}^{2}\right),\label{eq:guassian-sensor}
\end{equation}
where $\left\Vert z\right\Vert _{M}^{2}\triangleq z^{T}Mz$, and $H\in\mathbb{R}^{m\times n}$
is a fixed matrix of full row rank, which maps the external state
from $\mathbb{R}^{n}$ to ``sensory coordinates'' in $\mathbb{R}^{m}$.
Here, $h$ is the tuning curve height, i.e., the maximum firing rate
of the sensor; $\theta$ is the tuning curve center, i.e., the stimulus
value in sensory coordinates which elicits the highest firing rate;
and $R$ is the precision matrix of the Gaussian response curve in
the sensory space $\mathbb{R}^{m}$. Following neuroscience terminology,
we refer to the tuning curve center $\theta$ as the sensor's \emph{preferred
stimulus. }The inclusion of the matrix $H$ allows using high-dimensional
models where only some dimensions are observed, for example when the
full state includes velocities but only locations are directly observable.
In the one-dimensional case, $R^{-1/2}$ is the tuning curve width.

We consider several special forms of the population distribution $f$$\left(dh,d\theta,dR\right)$
where we can bring the approximate filter to closed form:

\subsubsection{A single sensor}

$f\left(dy\right)=\delta_{y}\left(dy\right)$, where $y=\left(h,\theta,R\right)$.

\subsubsection{Uniform population}

Here $h,R$ are fixed across all sensors, and $f\left(dh',d\theta,dR'\right)=\delta_{h}\left(dh'\right)\delta_{R}\left(dR'\right)d\theta$.

\subsubsection{Gaussian population}

Abusing notation slightly, we write $f\left(dh',d\theta,dR'\right)=\delta_{h}\left(dh'\right)\delta_{R}\left(dR'\right)f\left(d\theta\right)$
where

\begin{align}
f\left(d\theta\right) & =\mathcal{N}\left(\theta;c,\Sigma_{\mathrm{pop}}\right)d\theta\nonumber \\
 & =\left(2\pi\right)^{-n/2}\left|\Sigma_{\mathrm{pop}}\right|^{-1/2}\exp\left(-\frac{1}{2}\left\Vert \theta-c\right\Vert _{\Sigma_{\mathrm{pop}}^{-1}}^{2}\right)d\theta,\label{eq:gaussian-f}
\end{align}
for fixed $c\in\mathbb{R}^{m}$, and positive definite $\Sigma_{\mathrm{pop}}$.

We take $f$ to be normalized, since any scaling of $f$ may be included
in the coefficient $h$ in (\ref{eq:guassian-sensor}), resulting
in the same point-process. Thus, when used to approximate a large
population of sensors, the coefficient $h$ would be proportional
to the number of sensors.

\subsubsection{Uniform population on an interval}

In this case we assume a scalar state, $n=m=1$, and

\begin{align}
f\left(d\theta\right) & =1_{\left[a,b\right]}\left(\theta\right)d\theta,\label{eq:interval}
\end{align}
where similarly to the Gaussian population case, $h$ and $R$ are
fixed. Unlike the Gaussian case, here we find it more convenient not
to normalize the distribution.

\subsubsection{Finite mixtures}

$f\left(dy\right)=\sum_{i}\alpha_{i}f_{i}\left(dy\right),$ where
each $f_{i}$ is of one of the above forms. This form is quite general:
it includes populations where $\theta$ is distributed according to
a Gaussian mixture, as well as heterogeneous populations with finitely
many different values of $R$\@. The resulting filter derived below
includes a term for each component of the mixture.

\section{Assumed Density Filtering}

\subsection{Exact filtering equations}

Let $P\left(\cdot,t\right)$ denote the posterior density of $X_{t}$
given $\mathcal{N}_{t}$, and $\mathrm{E}_{P}^{t}\left[\cdot\right]$
the posterior expectation given $\mathcal{N}_{t}$. The prior density
$P\left(\cdot,0\right)$ is assumed to be known.

The problem of filtering a diffusion process $X$ from a doubly stochastic
Poisson process driven by $X$ is formally solved in \cite{Snyder1972}.
The result is extended to general marked point processes in the presence
of feedback in \cite{RhoSny1977}, where the authors derive a stochastic
PDE for the posterior density, which in our setting takes the form
\begin{align}
dP\left(x,t\right)= & \mathcal{L}_{t}^{*}P\left(x,t\right)dt+P\left(x,t^{-}\right)\int_{y\in\mathbf{Y}}\frac{\lambda_{t^{-}}\left(x;y\right)-\hat{\lambda}_{t^{-}}\left(y\right)}{\hat{\lambda}_{t^{-}}\left(y\right)}I\left(dt,dy\right),\label{eq:rhodes-snyder}
\end{align}
where $\mathcal{L}_{t}$ is the state's posterior infinitesimal generator
(Kolmogorov's backward operator), defined as $\mathcal{L}_{t}f\left(x\right)=\lim_{\Delta t\to0^{+}}\left(\mathrm{E}\left[f\left(X_{t+\Delta t}\right)|\mathcal{B}_{t}\right]-f\left(x\right)\right)/\Delta t$,
$\mathcal{L}_{t}^{*}$ is $\mathcal{L}_{t}$'s adjoint operator (Kolmogorov's
forward operator), and $I$ is defined in (\ref{eq:innovation}).
Expressions including $t^{-}$ are to be interpreted as left limits\footnote{The formulation in \cite{RhoSny1977} does not include left limits,
since it uses conditioning only on the strict past, resulting in a
left-continuous posterior. Our definition of $\mc N_{t}$ includes
the current time $t$, following the convention used in \cite{Segall1975-modelling}
and others, so that left limits are required at spike times.}. Note that the discrete part of the measure $I\left(dt,dy\right)$,
namely $N\left(dt,dy\right)$, contributes a discontinuous change
in the posterior at each spike time. Also note that in this closed-loop
setting, the infinitesimal generator is itself a random operator,
due to its dependence on past observations through the control law. 

The stochastic PDE (\ref{eq:rhodes-snyder}) is non-linear and non-local
(due to the dependence of $\hat{\lambda}_{t}\left(y\right)$ on $P\left(\cdot,t\right)$),
and therefore usually intractable. In \cite{RhoSny1977,Susemihl2014}
the authors consider linear dynamics with a Gaussian prior and Gaussian
sensors with centers distributed uniformly over the state space. In
this case, the posterior is Gaussian, and (\ref{eq:rhodes-snyder})
leads to closed-form ODEs for its moments. In our more general setting,
we can obtain exact equations for the posterior moments, as follows.

Let $\mu_{t}\triangleq\mathrm{E}_{P}^{t}X_{t},\tilde{X}_{t}\triangleq X_{t}-\mu_{t},\Sigma_{t}\triangleq\mathrm{E}_{P}^{t}[\tilde{X}_{t}\tilde{X}_{t}^{T}]$.
Using (\ref{eq:rhodes-snyder}), along with known results about the
form of the infinitesimal generator $\mathcal{L}_{t}$ for diffusion
processes (e.g. \cite{Oksendal2003}, Theorem 7.3.3), the first two
posterior moments can be shown to obey the following equations\footnote{see \cite{SusMeiOpp13} for derivation between spikes, or \cite{Segall1975-modelling}
for a derivation of (\ref{eq:mean}) via a different method}:
\begin{align}
d\mu_{t} & =\left(\mathrm{E}_{P}^{t}\left[A\left(X_{t}\right)\right]+B\left(U_{t}\right)\right)dt+\int_{\mathbf{Y}}\mathrm{E}_{P}^{t-}\left[\omega_{t^{-}}\left(y\right)X_{t^{-}}\right]I\left(dt,dy\right)\label{eq:mean}\\
d\Sigma_{t} & =\mathrm{E}_{P}^{t}\left[A\left(X_{t\left(y\right)}\right)\tilde{X}_{t}^{T}+\tilde{X}_{t}A\left(X_{t}\right)^{T}+D\left(X_{t}\right)D\left(X_{t}\right)^{T}\right]dt\nonumber \\
 & \quad+\int_{\mathbf{Y}}\mathrm{E}_{P}^{t^{-}}\left[\omega_{t^{-}}\left(y\right)\tilde{X}_{t^{-}}\tilde{X}_{t^{-}}^{T}\right]I\left(dt,dy\right)\nonumber \\
 & \quad-\int_{\mathbf{Y}}\mathrm{E}_{P}^{t^{-}}\left[\omega_{t^{-}}\left(y\right)X_{t^{-}}\right]\mathrm{E}_{P}^{t^{-}}\left[\omega_{t^{-}}\left(y\right)X_{t^{-}}^{T}\right]N\left(dt,dy\right)\label{eq:var}
\end{align}
where 
\[
\omega_{t}\left(y\right)\triangleq\frac{\lambda_{t}\left(X_{t};y\right)}{\hat{\lambda}_{t}\left(y\right)}-1,
\]
and similarly we write $\omega_{t}\left(x;y\right)\triangleq\lambda_{t}\left(x;y\right)/\hat{\lambda}_{t}\left(y\right)-1$.

In contrast with the more familiar LQG problem and the Kalman-Bucy
filter, here the posterior variance is random, and is generally not
monotonically decreasing even when estimating a constant state. However,
noting that $\E\left[I\left(dt,dy\right)\right]=0$, we may observe
from (\ref{eq:var}) that for a constant state ($A=D=0$), the \emph{expected}
posterior variance $\E\left[\Sigma_{t}\right]$ is decreasing.

Equations (\ref{eq:mean})-(\ref{eq:var}) are written in terms of
the innovation process $I\left(dt,dy\right)$ and the original point
process. Although this formulation highlights the relation to the
Kalman-Bucy filter, we will find it useful to rewrite them in a different
form, as follows,
\begin{align*}
d\mu_{t} & =d\mu_{t}^{\pi}+d\mu_{t}^{\mathrm{c}}+d\mu_{t}^{N},\\
d\Sigma_{t} & =d\Sigma_{t}^{\pi}+d\Sigma_{t}^{\mathrm{c}}+d\Sigma_{t}^{N},
\end{align*}
where $d\mu_{t}^{\pi},d\Sigma_{t}^{\pi}$ are the \emph{prior terms},\emph{
}corresponding to $\mathcal{L}_{t}^{*}P\left(x,t\right)$ in (\ref{eq:rhodes-snyder}),
and the remaining terms are divided into \emph{continuous update terms}
$d\mu_{t}^{\mathrm{c}},d\Sigma_{t}^{\mathrm{c}}$ (multiplying $dt$)
and \emph{discontinuous update terms} $d\mu_{t}^{N},d\Sigma_{t}^{N}$
(multiplying $N\left(dt,dy\right)$). Using (\ref{eq:innovation}),
we find
\begin{align}
d\mu_{t}^{\pi} & =\mathrm{E}_{P}^{t}\left[A\left(X_{t}\right)\right]dt+B\left(U_{t}\right)dt\label{eq:mean-p}\\
d\Sigma_{t}^{\pi} & =\mathrm{E}_{P}^{t}\left[A\left(X_{t}\right)\tilde{X}_{t}^{T}+\tilde{X}_{t}A\left(X_{t}\right)^{T}+D\left(X_{t}\right)D\left(X_{t}\right)^{T}\right]dt\label{eq:var-p}
\end{align}
\begin{align}
d\mu_{t}^{\mathrm{c}} & =-\int_{\mathbf{Y}}\mathrm{E}_{P}^{t}\left[\omega_{t}\left(y\right)X_{t}\right]\hat{\lambda}_{t}\left(y\right)f\left(dy\right)dt\label{eq:mean-c}\\
d\Sigma_{t}^{\mathrm{c}} & =-\int_{\mathbf{Y}}\mathrm{E}_{P}^{t}\left[\omega_{t}\left(y\right)\tilde{X}_{t}\tilde{X}_{t}^{T}\right]\hat{\lambda}_{t}\left(y\right)f\left(dy\right)dt\label{eq:var-c}
\end{align}
 
\begin{align}
d\mu_{t}^{N} & =\int_{\mathbf{Y}}\mathrm{E}_{P}^{t-}\left[\omega_{t^{-}}\left(y\right)X_{t^{-}}\right]N\left(dt,dy\right)\label{eq:mean-N}\\
d\Sigma_{t}^{N} & =\int_{\mathbf{Y}}\bigg(\mathrm{E}_{P}^{t^{-}}\left[\omega_{t^{-}}\left(y\right)\tilde{X}_{t^{-}}\tilde{X}_{t^{-}}^{T}\right]\label{eq:var-N}\\
 & \quad-\mathrm{E}_{P}^{t^{-}}\left[\omega_{t^{-}}\left(y\right)X_{t^{-}}\right]\mathrm{E}_{P}^{t^{-}}\left[\omega_{t^{-}}\left(y\right)X_{t^{-}}^{T}\right]\bigg)N\left(dt,dy\right).\nonumber 
\end{align}

The prior terms $d\mu_{t}^{\pi},d\Sigma_{t}^{\pi}$ represent the
known dynamics of $X$, and are the same terms appearing in the Kalman-Bucy
filter. These would be the only terms left if no measurements were
available, and would vanish for a static state. The continuous update
terms $d\mu_{t}^{\mathrm{c}},d\Sigma_{t}^{\mathrm{c}}$ represent
updates to the posterior between spikes that are not derived from
$X$'s dynamics, and therefore may be interpreted as corresponding
to information obtained from the absence of spikes. The discontinuous
update terms $d\mu_{t}^{N},d\Sigma_{t}^{N}$ contribute a change to
the posterior at spike times, depending on the spike's mark $y$,
and thus represent information obtained from the presence of a spike
as well as its associated mark.

Note that only the continuous update terms depend explicitly on the
population distribution $f$. Discontinuous updates depend on the
population distribution only indirectly through their influence on
the statistics of the point process $N$.

\subsection{ADF approximation}

While equations (\ref{eq:mean-p})-(\ref{eq:var-N}) are exact, they
are not practical, since they require computation of the full posterior
$\mathrm{E}_{P}^{t}\left[\cdot\right]$. To bring them to a closed
form, we use ADF with an assumed Gaussian density (see \cite{Opper98}
for details). Informally, this may be envisioned as integrating (\ref{eq:mean-p})-(\ref{eq:var-N})
while replacing the distribution $P$ by its approximating Gaussian
``at each time step''. The approximating Gaussian is obtained by
matching the first two moments of $P$ \cite{Opper98}. Note that
the solution of the resulting equations does not in general match
the first two moments of the exact solution, though it may approximate
it. Practically, the ADF approximation amounts to substituting the
normal distribution $\mathcal{N}(x;\mu_{t},\Sigma_{t})$ for $P(x,t)$
to compute the expectations in (\ref{eq:mean-p})-(\ref{eq:var-N}).

If the dynamics are linear, the prior updates (\ref{eq:mean-p})-(\ref{eq:var-p})
are easily computed in closed form after this substitution. Otherwise,
they may be approximated assuming the non-linear functions $A(x)$
and $D\left(x\right)D\left(x\right)^{T}$ may be written as power
series, using well-known results about the moments of Gaussian vectors.
The next sections are therefore devoted to the approximation of the
non-prior updates (\ref{eq:mean-c})-(\ref{eq:var-N}).

\subsection{Approximate form for Gaussian sensors}

We now proceed to apply the Gaussian ADF approximation $P\left(x,t\right)\approx\mathcal{N}\left(x;\mu_{t},\Sigma_{t}\right)$
to (\ref{eq:mean-c})-(\ref{eq:var-N}) in the case of Gaussian sensors
(\ref{eq:guassian-sensor}), deriving approximate filtering equations
written in terms of the population density $f\left(dh,d\theta,dR\right)$.
Abusing notation, from here on we use $\mu_{t},\Sigma_{t}$, and $P\left(x,t\right)$
to refer to the ADF approximation rather than to the exact values. 

To evaluate the posterior of expectations in (\ref{eq:mean-c})-(\ref{eq:var-N})
we first simplify the expression
\begin{equation}
P\left(x,t\right)\omega_{t}\left(x;y\right)=\frac{P\left(x,t\right)\lambda_{t}\left(x;y\right)}{\int P\left(\xi,y\right)\lambda_{t}\left(\xi;y\right)d\xi}-P\left(x,t\right).\label{eq:P-omega}
\end{equation}
Using the Gaussian ADF approximation $P\left(x,t\right)=\mathcal{N}\left(x;\mu_{t},\Sigma_{t}\right)$
and (\ref{eq:guassian-sensor}), we find 
\begin{align*}
P\left(x,t\right) & \lambda_{t}\left(x;h,\theta,R\right)=\\
 & =h\mathcal{N}\left(x;\mu_{t},\Sigma_{t}\right)\exp\left(-\frac{1}{2}\left\Vert Hx-\theta\right\Vert _{R}^{2}\right)\\
 & =\frac{h\exp\left(-\frac{1}{2}\left\Vert x-\mu_{t}\right\Vert _{\Sigma_{t}^{-1}}^{2}-\frac{1}{2}\left\Vert Hx-\theta\right\Vert _{R}^{2}\right)}{\sqrt{\left(2\pi\right)^{n}\left|\Sigma_{t}\right|}}\\
 & =\frac{h}{\sqrt{\left(2\pi\right)^{n}\left|\Sigma_{t}\right|}}\exp\left(-\frac{1}{2}\left\Vert H_{r}^{-1}\theta-\mu_{t}\right\Vert _{Q_{t}^{R}}^{2}-\frac{1}{2}\left\Vert x-\mu_{t}^{\theta}\right\Vert _{\Sigma_{t}^{-1}+H^{T}RH}^{2}\right),
\end{align*}
where $H_{r}^{-1}$ is any right inverse of $H$, and 
\begin{eqnarray*}
Q_{t}^{R} & \triangleq & \Sigma_{t}^{-1}\left(\Sigma_{t}^{-1}+H^{T}RH\right)^{-1}H^{T}RH,\\
\mu_{t}^{\theta} & \triangleq & \left(\Sigma_{t}^{-1}+H^{T}RH\right)^{-1}\left(Q_{t}\mu_{t}+H^{T}R\theta\right).
\end{eqnarray*}
An application of the Woodbury identity establishes the relation $Q_{t}^{R}=H^{T}S_{t}^{R}H$,
where 
\begin{equation}
S_{t}^{R}\triangleq\left(R^{-1}+H\Sigma_{t}H^{T}\right)^{-1},\label{eq:S}
\end{equation}
yielding 
\begin{multline}
P\left(x,t\right)\lambda_{t}\left(x;h,\theta,R\right)\\
\quad=\frac{h}{\sqrt{\left(2\pi\right)^{n}\left|\Sigma_{t}\right|}}\exp\left(-\frac{1}{2}\left\Vert \theta-H\mu_{t}\right\Vert _{S_{t}^{R}}^{2}-\frac{1}{2}\left\Vert x-\mu_{t}^{\theta}\right\Vert _{\Sigma_{t}^{-1}+H^{T}RH}^{2}\right),\label{eq:post-times-rate}
\end{multline}
and by normalizing this Gaussian (see (\ref{eq:P-omega})) we find
that 
\begin{align*}
P\left(x,t\right)\omega_{t}\left(x;h,\theta,R\right) & =\mathcal{N}\left(x;\mu_{t}^{\theta},\left(\Sigma_{t}^{-1}+H^{T}RH\right)^{-1}\right)-P\left(x,t\right)
\end{align*}
yielding
\begin{align*}
\mathrm{E}_{P}^{t}\left[\omega_{t}\left(h,\theta,R\right)X_{t}\right] & =\mu_{t}^{\theta}-\mu_{t}\\
\mathrm{E}_{P}^{t}\left[\omega_{t}\left(h,\theta,R\right)\tilde{X}_{t}\tilde{X}_{t}^{T}\right] & =\left(\Sigma_{t}^{-1}+H^{T}RH\right)^{-1}+\left(\mu_{t}-\mu_{t}^{\theta}\right)\left(\mu_{t}-\mu_{t}^{\theta}\right)^{T}-\Sigma_{t}.
\end{align*}
Substituting the definition of $\mu_{t}^{\theta}$ and simplifying
using the Woodbury identity yields
\begin{align*}
\mathrm{E}_{P}^{t} & \left[\omega_{t}\left(h,\theta,R\right)X_{t}\right]=-\Sigma_{t}H^{T}S_{t}^{R}\left(H\mu_{t}-\theta\right),\\
\mathrm{E}_{P}^{t} & \left[\omega_{t}\left(h,\theta,R\right)\tilde{X}_{t}\tilde{X}_{t}^{T}\right]=\Sigma_{t}H^{T}\left(S_{t}^{R}\left(H\mu_{t}-\theta\right)\left(H\mu_{t}-\theta\right)^{T}S_{t}^{R}-S_{t}^{R}\right)H\Sigma_{t}.
\end{align*}
Plugging this result into (\ref{eq:mean-c})-(\ref{eq:var-N}) yields

\begin{align}
d\mu_{t}^{\mathrm{c}} & =\Sigma_{t}H^{T}\int_{\mathbf{Y}}S_{t}^{R}\left(H\mu_{t}-\theta\right)\hat{\lambda}_{t}\left(h,\theta,R\right)f\left(dh,d\theta,dR\right)dt\label{eq:mean-c-gaussian}\\
d\Sigma_{t}^{\mathrm{c}} & =\Sigma_{t}H^{T}\int_{\mathbf{Y}}\left(S_{t}^{R}-S_{t}^{R}\left(H\mu_{t}-\theta\right)\left(H\mu_{t}-\theta\right)^{T}S_{t}^{R}\right)\nonumber \\
 & \qquad\qquad\qquad\times\hat{\lambda}_{t}\left(h,\theta,R\right)f\left(dh,d\theta,dR\right)H\Sigma_{t}dt\label{eq:var-c-gaussian}\\
d\mu_{t}^{N} & =\Sigma_{t^{-}}H^{T}\int_{\mathbf{Y}}S_{t^{-}}^{R}\left(\theta-H\mu_{t^{-}}\right)N\left(dt,d\left[h,\theta,R\right]\right)\label{eq:mean-N-guassian}\\
d\Sigma_{t}^{N} & =-\Sigma_{t^{-}}H^{T}\int_{\mathbf{Y}}S_{t^{-}}^{R}N\left(dt,d\left[h,\theta,R\right]\right)H\Sigma_{t^{-}}\label{eq:var-N-gaussian}
\end{align}

We also note that integrating (\ref{eq:post-times-rate}) over $x$
yields
\begin{align}
\hat{\lambda}_{t}\left(h,\theta,R\right) & =h\sqrt{\frac{\left|S_{t}^{R}\right|}{\left|R\right|}}\exp\left(-\frac{1}{2}\left\Vert \theta-H\mu_{t}\right\Vert _{S_{t}^{R}}^{2}\right),\label{eq:self-rate-gaussian}
\end{align}
where we computed the coefficient using the equality $\left|I+\Sigma_{t}H^{T}RH\right|=\left|R\right|/\left|S_{t}^{R}\right|$,
derived by application of Sylvester's determinant identity. 

To gain some insight into these equations, consider the case $H=I$.
As seen from the discontinuous update equations (\ref{eq:mean-N-guassian})-(\ref{eq:var-N-gaussian}),
when a spike with mark $\left(h,\theta,R\right)$ occurs, the posterior
mean moves towards the location mark $\theta$, and the posterior
variance decreases (in the sense that $\Sigma_{t^{+}}-\Sigma_{t^{-}}$
is negative definite). Neither update depends on $h$.

In the scalar case $m=n=1$, with $H=1$, $\sigma_{t}^{2}=\Sigma_{t},\sigma_{\mathrm{r}}^{2}=R^{-1},a=A,d=D$,
the discontinuous update equations (\ref{eq:mean-N-guassian})-(\ref{eq:var-N-gaussian})
read 
\begin{align}
d\mu_{t}^{N} & =\int_{\mathbf{Y}}\frac{\sigma_{t^{-}}^{2}}{\sigma_{t^{-}}^{2}+\sigma_{\mathrm{r}}^{2}}\left(\theta-\mu_{t^{-}}\right)N\left(dt,d\left[h,\theta,\sigma_{r}^{2}\right]\right)\label{eq:mean-N-guassian-1d}\\
d\Sigma_{t}^{N} & =-\int_{\mathbf{Y}}\frac{\sigma_{t^{-}}^{2}}{\sigma_{t^{-}}^{2}+\sigma_{\mathrm{r}}^{2}}\sigma_{t^{-}}^{2}N\left(dt,d\left[h,\theta,\sigma_{r}^{2}\right]\right)\label{eq:var-N-gaussian-1d}
\end{align}

The continuous mean update equation (\ref{eq:mean-c-gaussian}) also
admits an intuitive interpretation, in the case where all sensors
share the same precision matrix $R$, i.e. $f\left(d\left[h,\theta,R'\right]\right)=f\left(dh,d\theta\right)\delta_{R}\left(dR'\right)$.
In this case, the equation reads
\begin{align*}
d\mu_{t}^{\mathrm{c}} & =\Sigma_{t}H^{T}S_{t}^{R}\left(H\mu_{t}-\int_{\mathbf{Y}}\theta\nu_{t}\left(\theta\right)f\left(dh,d\theta\right)\right)\hat{\lambda}_{t}^{f}dt
\end{align*}
where $\nu_{t}\left(\theta\right)\triangleq\hat{\lambda}_{t}\left(h,\theta,R\right)/\hat{\lambda}_{t}^{f}$
(which does not depend on $h$). The normalized intensity kernel $\nu_{t}\left(\theta\right)f\left(dh,d\theta\right)$
may be interpreted heuristically as the distribution of the mark $\left(h,\theta\right)$
of the next event provided it occurs immediately. Thus, the absence
of events drives the posterior mean away from the expected location
of the next mark. The strength of this effect scales with $\hat{\lambda}_{t}^{f}$,
which is the rate of the unmarked process $N_{t}$ w.r.t.~its natural
filtration, i.e., the total expected rate of spikes given the firing
history. This behavior is qualitatively similar to the result obtained
in \cite{BobMeiEld09} for a finite population of sensors observing
a continuous-time finite-state Markov process, where the posterior
probability between spikes concentrates on states with lower total
firing rate.

\subsection{Closed form approximations in special cases}

Using (\ref{eq:self-rate-gaussian}), we now evaluate the continuous
update equations (\ref{eq:mean-c-gaussian})-(\ref{eq:var-c-gaussian})
for the specific forms of the population distribution $f\left(dy\right)$
listed in section \ref{sub:special-cases}. Note that the discontinuous
update equations (\ref{eq:mean-N-guassian})-(\ref{eq:var-N-gaussian})
do not depend on the population distribution $f$, and are already
in closed form.

\subsubsection{Single sensor}

The result for a single sensor with parameters $h,\theta,R$ is trivial
to obtain from (\ref{eq:mean-c-gaussian})-(\ref{eq:var-c-gaussian}),
yielding
\begin{align}
d\mu_{t}^{\mathrm{c}} & =\Sigma_{t}H^{T}S_{t}^{R}\left(H\mu_{t}-\theta\right)\hat{\lambda}_{t}^{f}dt,\label{eq:mean-c-single}\\
d\Sigma_{t}^{\mathrm{c}} & =\Sigma_{t^{-}}H^{T}\left(S_{t}^{R}-S_{t}^{R}\left(H\mu_{t}-\theta\right)\left(H\mu_{t}-\theta\right)^{T}S_{t}^{R}\right)\hat{\lambda}_{t}^{f}H\Sigma_{t^{-}}dt,\label{eq:var-c-single}
\end{align}
where $\hat{\lambda}_{t}^{f}=\hat{\lambda}_{t}\left(h,\theta,R\right)$
as given by (\ref{eq:self-rate-gaussian}), and $S_{t}^{R}$ is defined
in (\ref{eq:S}).

\subsubsection{Uniform population}

Here all sensors share the same height $h$ and precision $R$, whereas
the location parameter $\theta$ covers $\mathbb{R}^{m}$ uniformly,
i.e. $f\left(d\theta\right)=d\theta$. A straightforward calculation
from (\ref{eq:mean-c-gaussian})-(\ref{eq:var-c-gaussian}) and (\ref{eq:self-rate-gaussian})
yields
\begin{align}
d\mu_{t}^{\mathrm{c}} & =0,\quad d\Sigma_{t}^{\mathrm{c}}=0,\label{eq:mean-var-uniform}
\end{align}
in agreement with the (exact) result obtained in \cite{RhoSny1977},
where the filtering equations only include the prior term and the
discontinuous update term.

\subsubsection{Gaussian population}

Here $R,h$ are constant, and $f\left(d\theta\right)$ is given in
(\ref{eq:gaussian-f}). Using (\ref{eq:self-rate-gaussian}),
\begin{align*}
\hat{\lambda}_{t}\left(h,\theta,R\right)f\left(d\theta\right) & =h\sqrt{\frac{\left|S_{t}^{R}\right|}{\left|R\right|}}\exp\left(-\frac{1}{2}\left\Vert \theta-H\mu_{t}\right\Vert _{S_{t}^{R}}^{2}\right)\mathcal{N}\left(\theta;c,\Sigma_{\mathrm{pop}}\right)d\theta.
\end{align*}
An analogous computation to the derivation of (\ref{eq:post-times-rate})
and (\ref{eq:self-rate-gaussian}) above yields
\begin{align}
\hat{\lambda}_{t}\left(h,\theta,R\right)f\left(d\theta\right) & =\hat{\lambda}_{t}^{f}\cdot\mathcal{N}\left(\theta;\mu_{t}^{f},\left(\Sigma_{\mathrm{pop}}^{-1}+S_{t}^{R}\right)^{-1}\right)d\theta,\nonumber \\
\hat{\lambda}_{t}^{f} & =\int\hat{\lambda}_{t}\left(h,\theta,R\right)f\left(d\theta\right)\nonumber \\
 & =h\sqrt{\frac{\left|Z_{t}^{R}\right|}{\left|R\right|}}\exp\left(-\frac{1}{2}\left\Vert c-H\mu_{t}\right\Vert _{Z_{t}^{R}}^{2}\right)\label{eq:total-rate-gauss-gauss}
\end{align}
where
\begin{align}
Z_{t}^{R} & \triangleq\left(\Sigma_{\mathrm{pop}}+\left(S_{t}^{R}\right)^{-1}\right)^{-1}=\left(\Sigma_{\mathrm{pop}}+R^{-1}+H\Sigma_{t}H^{T}\right)^{-1}\label{eq:Z}\\
\mu_{t}^{f} & \triangleq\left(\Sigma_{\mathrm{pop}}^{-1}+S_{t}^{R}\right)^{-1}\left(S_{t}^{R}\mu_{t}+\Sigma_{\mathrm{pop}}^{-1}c\right).\nonumber 
\end{align}
Substituting into (\ref{eq:mean-c-gaussian})-(\ref{eq:var-c-gaussian})
and simplifying yields the following continuous update equations 
\begin{align}
d\mu_{t}^{\mathrm{c}} & =\Sigma_{t}H^{T}Z_{t}^{R}\left(H\mu_{t}-c\right)\hat{\lambda}_{t}^{f}dt\label{eq:mean-c-gauss-gauss}\\
d\Sigma_{t}^{\mathrm{c}} & =\Sigma_{t}H^{T}\Big(Z_{t}^{R}-Z_{t}^{R}\left(H\mu_{t}-c\right)\left(H\mu_{t}-c\right)^{T}Z_{t}^{R}\Big)H\Sigma_{t}\hat{\lambda}_{t}^{f}dt,\label{eq:var-c-gauss-gauss}
\end{align}
where $Z_{t}^{R}$ and $\hat{\lambda}_{t}^{f}$ are given by (\ref{eq:Z})
and (\ref{eq:total-rate-gauss-gauss}), respectively. These updates
generalize the single-sensor updates (\ref{eq:mean-c-single})-(\ref{eq:var-c-single}),
with the population center $c$ taking the place of the single location
parameter, and $Z_{t}^{R}$ substituting $S_{t}^{R}$. The single
sensor case is obtained when $\Sigma_{\mathrm{pop}}=0$.

It is illustrative to consider these equations in the scalar case
$m=n=1$, with $H=1$. Letting $\sigma_{t}^{2}=\Sigma_{t},\sigma_{\mathrm{r}}^{2}=R^{-1},\sigma_{\mathrm{pop}}^{2}=\Sigma_{\mathrm{pop}}$
yields
\begin{align}
d\mu_{t}^{\mathrm{c}} & =\frac{\sigma_{t}^{2}}{\sigma_{t}^{2}+\sigma_{\mathrm{r}}^{2}+\sigma_{\mathrm{pop}}^{2}}\left(\mu_{t}-c\right)\hat{\lambda}_{t}^{f}dt,\label{eq:mean-c-gauss-gauss-1d}\\
d\sigma_{t}^{2,\mathrm{c}} & =\frac{\sigma_{t}^{2}}{\sigma_{t}^{2}+\sigma_{\mathrm{r}}^{2}+\sigma_{\mathrm{pop}}^{2}}\left(1-\frac{\left(\mu_{t}-c\right)^{2}}{\sigma_{t}^{2}+\sigma_{\mathrm{r}}^{2}+\sigma_{\mathrm{pop}}^{2}}\right)\sigma_{t}^{2}\hat{\lambda}_{t}^{f}dt,\label{eq:var-c-gauss-gauss-1d}
\end{align}
where 
\[
\hat{\lambda}_{t}^{f}=h\sqrt{2\pi\sigma_{\mathrm{r}}^{2}}\mathcal{N}\left(\mu_{t};c,\sigma_{t}^{2}+\sigma_{\mathrm{r}}^{2}+\sigma_{\mathrm{pop}}^{2}\right).
\]

\begin{figure*}
\subfloat[Gaussian population ((\ref{eq:mean-c-gauss-gauss-1d})-(\ref{eq:var-c-gauss-gauss-1d})).
Parameters are $c=0,\sigma_{\mathrm{r}}^{2}=0.25,h=1,\sigma_{t}^{2}=1$.
The case $\sigma_{\mathrm{pop}}^{2}=0$ corresponds to a single sensor.]{\includegraphics[bb=0bp 0bp 534bp 390bp,width=0.45\textwidth]{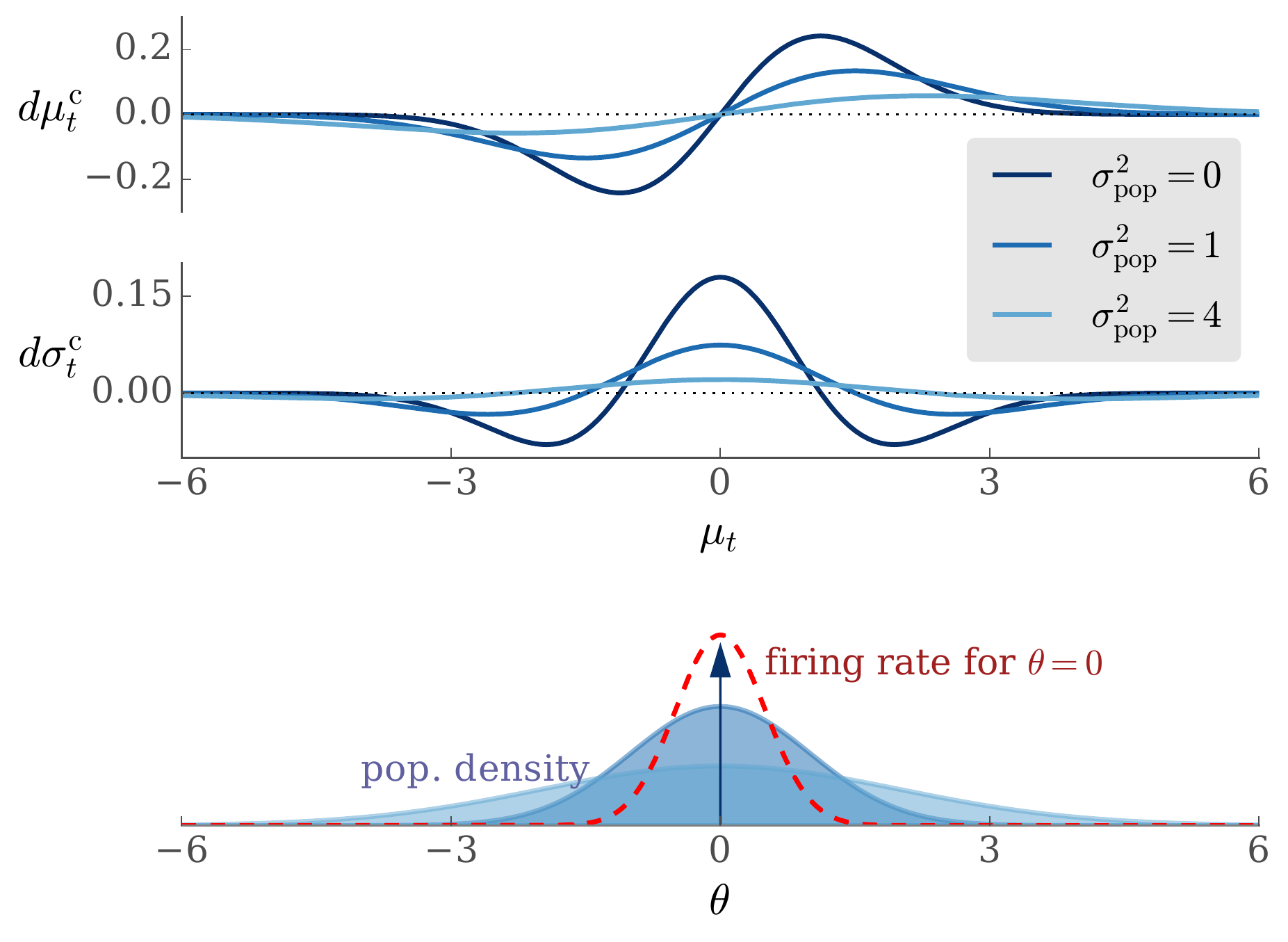}\label{gaussian}}\hfill{}\subfloat[{Uniform population on an interval ((\ref{eq:mean-c-interval})-(\ref{eq:var-c-interval})).
Parameters are $\left[a,b\right]=\left[-1,1\right],h=1,\sigma_{t}^{2}=0.01$.}]{\includegraphics[bb=0bp 0bp 534bp 390bp,width=0.45\textwidth]{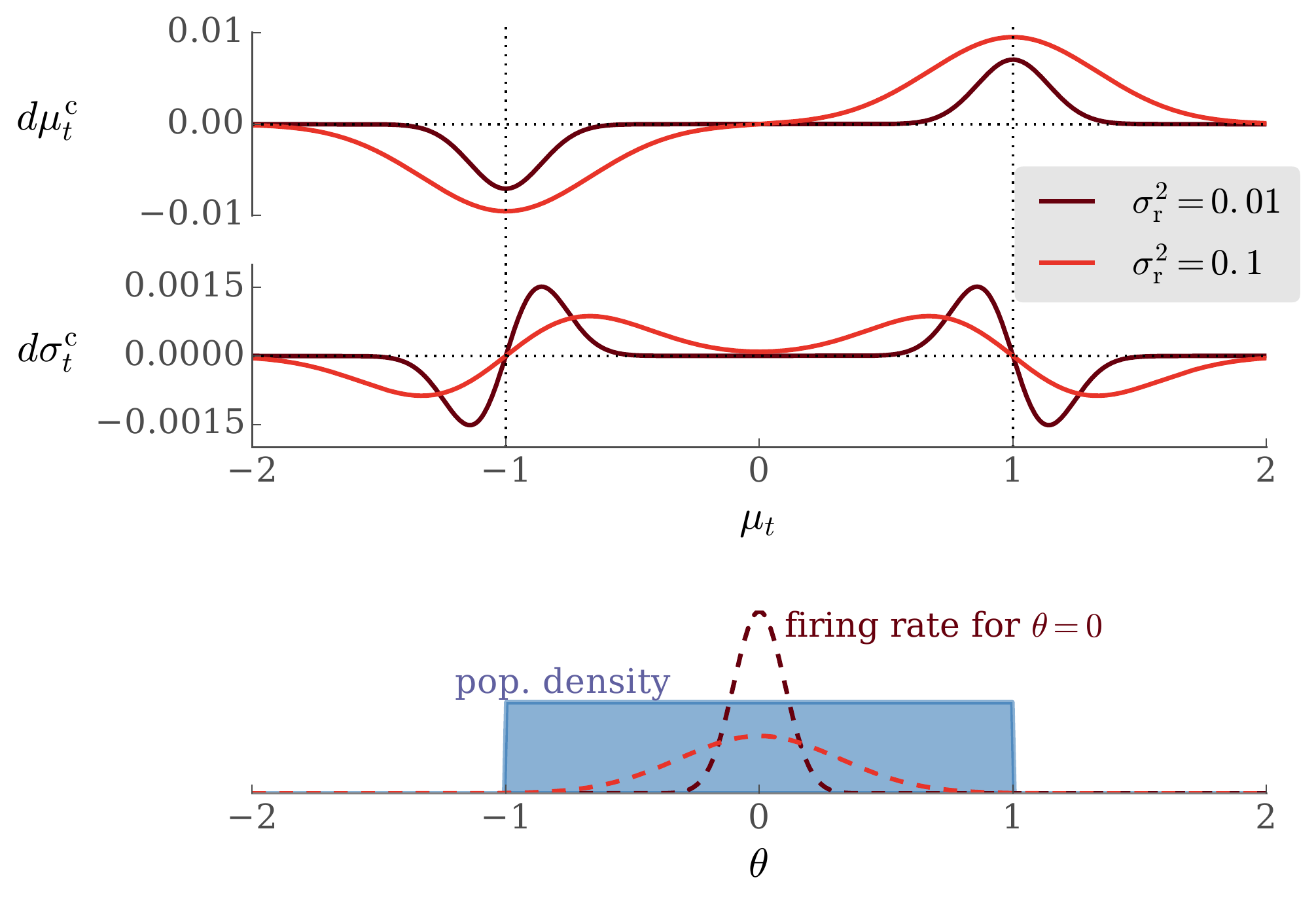}

\label{interval}}\caption{Continuous update terms as a function of the current posterior mean
estimate, for a 1-d state observed through a population of Gaussian
sensors (\ref{eq:guassian-sensor}). The population density is Gaussian
on the left plot, and uniform on the interval $\left[-1,1\right]$
on the right plot. The bottom plots shows the population density $f\left(d\theta\right)/d\theta$
and a tuning curve $\lambda_{t}\left(x;\theta\right)$ for $\theta=0$.}
\label{continuous-terms}
\end{figure*}

Figure \ref{gaussian} demonstrates the continuous update terms (\ref{eq:mean-c-gauss-gauss-1d})-(\ref{eq:var-c-gauss-gauss-1d})
as a function of the current mean estimate $\mu_{t}$, for various
values of the population variance $\sigma_{\mathrm{pop}}^{2}$, including
the case of a single sensor, $\sigma_{\mathrm{pop}}^{2}=0$. The continuous
update term $d\mu_{t}^{\mathrm{c}}$ pushes the posterior mean $\mu_{t}$
away from the the population center $c$ in the absence of spikes.
This effect weakens as $\left|\mu_{t}-c\right|$ grows due to the
factor $\hat{\lambda}_{t}^{f}$, consistent with the idea that far
from $c$, the lack of events is less surprising, hence less informative.
The continuous variance update term $d\sigma_{t}^{2,\mathrm{c}}$
increases the variance when $\mu_{t}$ is near $\theta$, otherwise
decreases it. This stands in contrast with the Kalman-Bucy filter,
where the posterior variance cannot increase when estimating a static
state.

\subsubsection{Uniform population on an interval}

In this case $R,h$ are constant, and $f\left(d\theta\right)$ is
given in (\ref{eq:interval}). Here we assume $m=n=1$ and $H=1$,
so (\ref{eq:mean-c-gaussian})-(\ref{eq:var-c-gaussian}) take the
form
\begin{align*}
d\mu_{t}^{\mathrm{c}} & =\frac{\sigma_{t}^{2}}{\sigma_{t}^{2}+\sigma_{\mathrm{r}}^{2}}\int_{\theta\in\mathbb{R}}\left(\mu_{t}-\theta\right)\hat{\lambda}_{t}\left(\theta\right)f\left(d\theta\right)dt\\
d\sigma_{t}^{2,\mathrm{c}} & =\left(\hat{\lambda}_{t}^{f}-\frac{\int\left(\theta-\mu_{t}\right)^{2}\hat{\lambda}_{t}\left(\theta\right)f\left(d\theta\right)}{\sigma_{t}^{2}+\sigma_{\mathrm{r}}^{2}}\right)\frac{\sigma_{t}^{2}}{\sigma_{t}^{2}+\sigma_{\mathrm{r}}^{2}}\sigma_{t}^{2}dt\\
\hat{\lambda}_{t}\left(\theta\right) & =h\sqrt{2\pi\sigma_{\mathrm{r}}^{2}}\mathcal{N}\left(\theta;\mu_{t},\sigma_{\mathrm{r}}^{2}+\sigma_{t}^{2}\right),
\end{align*}
where $\sigma_{t}^{2}=\Sigma_{t},\sigma_{\mathrm{r}}^{2}=R^{-1}$,
and we suppressed the dependence of $\hat{\lambda}$ on $h,R$ from
the notation, since $h,R$ are fixed. Let $\phi\left(x\right)=\mathcal{N}\left(x;0,1\right),\Phi\left(x\right)=\int_{-\infty}^{x}\phi$.
A straightforward computation using the moments of the truncated Gaussian
distribution yields

\begin{align*}
\hat{\lambda}_{t}^{f}=\int_{a}^{b}\hat{\lambda}_{t}\left(\theta\right)d\theta & =h\sqrt{2\pi\sigma_{\mathrm{r}}^{2}}Z_{t},\\
\int\left(\theta-\mu_{t}\right)\hat{\lambda}_{t}\left(\theta\right)f\left(d\theta\right) & =-h\sqrt{2\pi\sigma_{\mathrm{r}}^{2}}z_{t}\sqrt{\sigma_{t}^{2}+\sigma_{\mathrm{r}}^{2}},\\
\int\left(\theta-\mu_{t}\right)^{2}\hat{\lambda}_{t}\left(\theta\right)f\left(d\theta\right) & =\left[\hat{\lambda}_{t}^{f}-h\sqrt{2\pi\sigma_{\mathrm{r}}^{2}}z'_{t}\right]\left(\sigma_{t}^{2}+\sigma_{\mathrm{r}}^{2}\right),
\end{align*}
where
\begin{align*}
\alpha_{t} & =\frac{a-\mu_{t}}{\sqrt{\sigma_{t}^{2}+\sigma_{\mathrm{r}}^{2}}}\,, & Z_{t} & =\Phi\left(\beta_{t}\right)-\Phi\left(\alpha_{t}\right),\\
\beta_{t} & =\frac{b-\mu_{t}}{\sqrt{\sigma_{t}^{2}+\sigma_{\mathrm{r}}^{2}}}\,, & z_{t} & =\phi\left(\beta_{t}\right)-\phi\left(\alpha_{t}\right),\\
 &  & z'_{t} & =\beta_{t}\phi\left(\beta_{t}\right)-\alpha_{t}\phi\left(\alpha_{t}\right),
\end{align*}
yielding
\begin{align}
d\mu_{t}^{\mathrm{c}} & =h\sqrt{2\pi\sigma_{\mathrm{r}}^{2}}\sqrt{\frac{\sigma_{t}^{2}}{\sigma_{t}^{2}+\sigma_{\mathrm{r}}^{2}}}z_{t}\sigma_{t}dt\label{eq:mean-c-interval}\\
d\sigma_{t}^{2,\mathrm{c}} & =h\sqrt{2\pi\sigma_{\mathrm{r}}^{2}}\frac{\sigma_{t}^{2}}{\sigma_{t}^{2}+\sigma_{\mathrm{r}}^{2}}z'_{t}\sigma_{t}^{2}dt\label{eq:var-c-interval}
\end{align}

Figure \ref{interval} demonstrates the continuous update terms (\ref{eq:mean-c-interval})-(\ref{eq:var-c-interval})
as a function of the current mean estimate $\mu_{t}$. When the mean
estimate is around an endpoint of the interval, the mean update $\mu_{t}^{\mathrm{c}}$
pushes the posterior mean outside the interval in the absence of spikes.
The posterior variance $\sigma_{t}^{2}$ decreases outside the interval,
where the absence of spikes is expected, and increases inside the
interval, where it is unexpected\footnote{This holds only approximately, when the tuning curve width is not
too large relative to the size of the interval. For wider tuning curves
the behavior becomes similar to the single sensor case.}. When the posterior mean is not near the interval endpoints, the
updates are near zero, consistently with the uniform population case
(\ref{eq:mean-var-uniform}).

\subsubsection{Finite Mixtures}

Assume $f$ is any finite mixture of measures 
\[
f\left(dy\right)=\sum_{i}\alpha_{i}f_{i}\left(dy\right),
\]
where $f_{i}$ is of one of the above forms. We note that the continuous
updates (\ref{eq:mean-c})-(\ref{eq:var-c}) are linear in $f$, so
they are obtained by the appropriate weighted sums of the filters
derived above for the various special forms of $f_{i}$.

In particular, for a general mixture of the first three forms\footnote{In the one-dimensional case, the fourth form may be included similarly},
\begin{align*}
f\left(dh,dR,d\theta\right) & =\sum_{i}\alpha_{i}^{\delta}\delta_{h_{i}}\left(dh\right)\delta_{R_{i}}\left(dR\right)\delta_{\theta_{i}}\left(d\theta\right)\\
 & \quad+\sum_{i}\alpha_{i}^{U}\delta_{h_{i}^{U}}\left(dh\right)\delta_{R_{i}^{U}}\left(dR\right)d\theta\\
 & \quad+\sum_{i}\alpha_{i}^{\mc N}\delta_{h_{i}^{\mc N}}\left(dh\right)\delta_{R_{i}^{\mc N}}\left(dR\right)\mathcal{N}\left(\theta;c_{i},\Sigma_{\mathrm{pop}}^{\left(i\right)}\right)d\theta,
\end{align*}
the resulting continuous update terms are given by
\begin{align*}
d\mu_{t}^{\mathrm{c}}=\Sigma_{t}H^{T}\Bigg( & \sum_{i}\alpha_{i}^{\delta}S_{t}^{i}\left(H\mu_{t}-\theta_{i}\right)\hat{\lambda}_{t}\left(h_{i},\theta_{i},R_{i}\right)\\
\qquad+ & \sum_{i}\alpha_{i}^{\mc N}Z_{t}^{i}\left(H\mu_{t}-c_{i}\right)\hat{\lambda}_{t}^{f,i}\bigg)dt,
\end{align*}
\begin{align*}
d\Sigma_{t}^{\mathrm{c}}=\Sigma_{t}H^{T}\Bigg\{ & \sum_{i}\alpha_{i}^{\delta}\left[S_{t}^{i}-S_{t}^{i}\left(H\mu_{t}-\theta_{i}\right)\left(H\mu_{t}-\theta_{i}\right)^{T}S_{t}^{i}\right]\hat{\lambda}_{t}\left(h_{i},\theta_{i},R_{i}\right)\\
 & +\sum_{i}\alpha_{i}^{\mc N}\left[Z_{t}^{i}-Z_{t}^{i}\left(H\mu_{t}-c_{i}\right)\left(H\mu_{t}-c_{i}\right)^{T}Z_{t}^{i}\right]\hat{\lambda}_{t}^{f,i}\Bigg\} H\Sigma_{t}dt,
\end{align*}
where $\hat{\lambda}_{t}\left(h,\theta,R\right)$ is given by (\ref{eq:self-rate-gaussian}),
$S_{t}^{i}=S_{t}^{R_{i}},Z_{t}^{i}=Z_{t}^{R_{i}^{\mc N}}$ as defined
in (\ref{eq:S}),(\ref{eq:Z}), and similarly,
\[
\hat{\lambda}_{t}^{f,i}=h_{i}^{\mc N}\sqrt{\frac{\left|Z_{t}^{i}\right|}{\left|R_{i}^{\mc N}\right|}}\exp\left(-\frac{1}{2}\left\Vert c_{i}-H\mu_{t}\right\Vert _{Z_{t}^{i}}^{2}\right)
\]
(cf. (\ref{eq:total-rate-gauss-gauss})). The uniform components of
the mixture do not appear explicitly in the filter, since the continuous
update part vanishes for the uniform case (see (\ref{eq:mean-var-uniform})),
though they would influence the discontinuous update terms through
their effect on the statistics of $N$.

\section{Evaluation of filter}

\subsection{Examples and comparison to particle filter}

\begin{figure*}
\subfloat[high rate of events: $h=1000$]{\includegraphics[bb=0bp 0bp 534bp 390bp,width=0.45\textwidth]{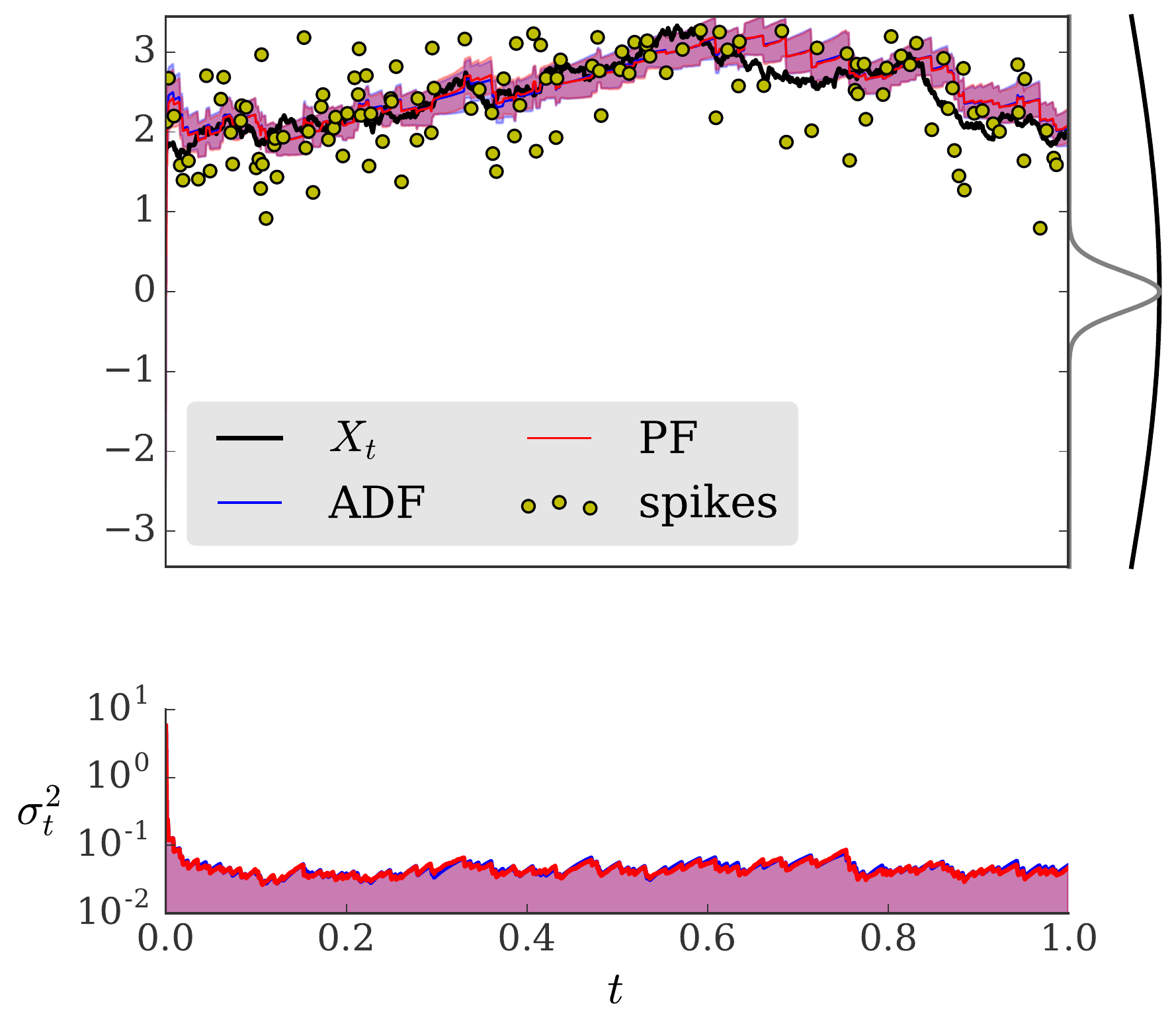}\label{high-rate}}\hfill{}\subfloat[low rate of events: $h=2$]{\includegraphics[bb=0bp 0bp 534bp 390bp,width=0.45\textwidth]{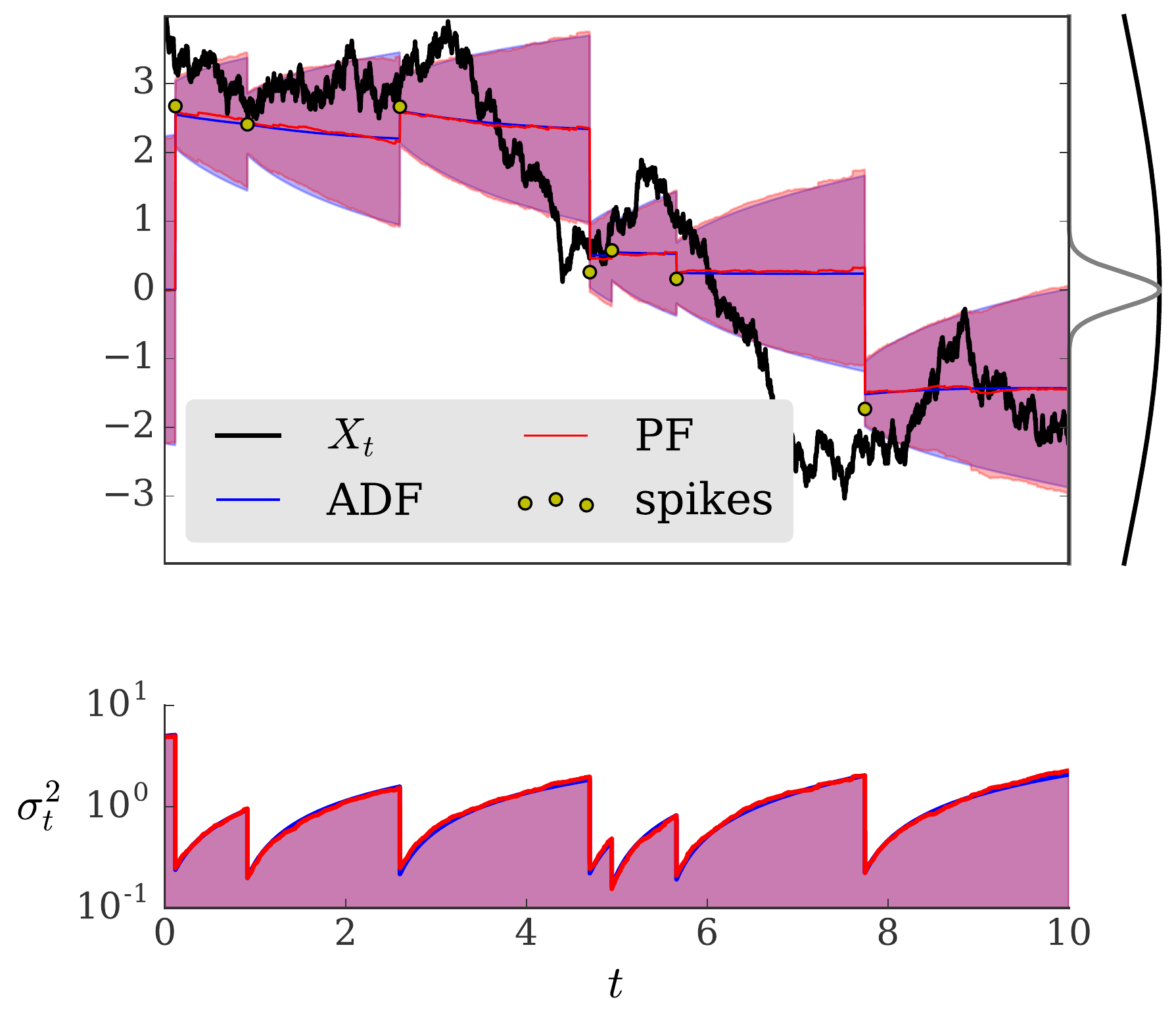}\label{low-rate}}\caption{Two examples of a linear one-dimensional process observed through
a Gaussian population (\ref{eq:gaussian-f}) of Gaussian sensors (\ref{eq:guassian-sensor}),
filtered using the ADF approximation (namely, equations (\ref{eq:mean-p})-(\ref{eq:var-p}),(\ref{eq:mean-c-gaussian})-(\ref{eq:var-c-gaussian}),(\ref{eq:mean-c-gauss-gauss})-(\ref{eq:var-c-gauss-gauss})),
and using a particle filter. Each dot correspond to a spike with the
vertical location indicating the sensor's location parameter $\theta$.
The approximate posterior means obtained from ADF and particle filtering
are shown in blue and red respectively, with the corresponding posterior
standard deviations in shaded areas of the respective colors. The
curves to the right of the graph show the preferred stimulus density
(black), and a sensor's response rate function (\ref{eq:guassian-sensor})
centered at $\theta=0$ (gray), arbitrarily normalized to the same
height for visualization. The bottom graph shows the posterior variance.
Parameters used in both examples: $a=-0.1,d=1,H=1,\sigma_{\mathrm{pop}}^{2}=4,R^{-1}=0.25,c=0,\mu_{0}=0,\sigma_{0}^{2}=1$
(note the different extent of the time axis). The observed processes
were initialized from their steady-state distribution. The dynamics
were discretized with time step $\Delta t=10^{-3}$. The particle
filter uses 1000 particles with systematic resampling (see, e.g.,
\cite{DouJoh09}) at each time step.}

\label{1d-filtering}
\end{figure*}

Since the filter (\ref{eq:mean-p})-(\ref{eq:var-N}) is based on
an assumed density approximation, its results may be inexact. We tested
the accuracy of the filter in the Gaussian population case (\ref{eq:mean-c-gauss-gauss})-(\ref{eq:var-c-gauss-gauss}),
by numerical comparison with Particle Filtering (PF) \cite{DouJoh09}.

Figure \ref{1d-filtering} shows two examples of filtering a one-dimensional
process observed through a Gaussian population of Gaussian sensors
(\ref{eq:gaussian-f}), using both the ADF approximation (\ref{eq:mean-c-gauss-gauss-1d})-(\ref{eq:var-c-gauss-gauss-1d})
and a Particle Filter (PF) for comparison. See the figure caption
for precise details.

\begin{figure}
\includegraphics[width=1\columnwidth]{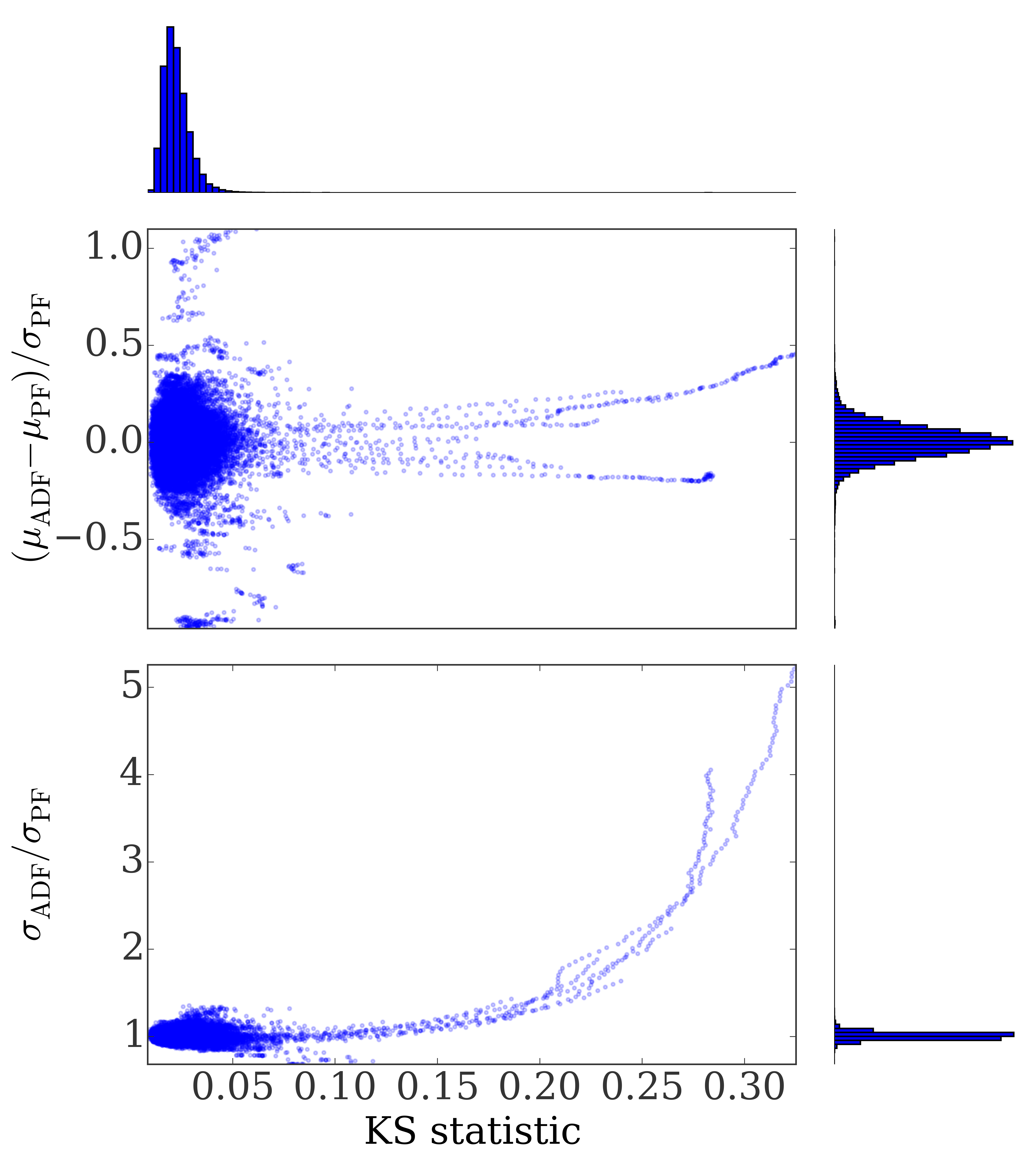}\caption{Approximation errors (relative to particle filter) vs. KS statistic
of particle distribution. The KS statistic is plotted against the
estimation errors $\epsilon_{\mu},\epsilon_{\sigma}$ (see main text).
Each point corresponds to a single simulation time step. The histograms
show the distribution of the KS statistic, of $\epsilon_{\mu}$ and
of $\epsilon_{\sigma}$. Results obtained from 100 trials of length
$T=1$, with parameters as in Figure \ref{high-rate}. }
\label{ks}
\end{figure}

Figure \ref{ks} shows the distribution of approximation errors and
their relation to the deviation of the posterior from Gaussian. The
deviation of the particle distribution from Gaussian is quantified
using the Kolmogorov-Smirnov (KS) statistic $\sup_{x}\left|F\left(x\right)-G\left(x\right)\right|$
where $F$ is the particle distribution cdf and $G$ is the cdf of
a Gaussian matching $F$'s first two moments. The approximation errors
plotted are the relative error in the mean estimate $\epsilon_{\mu}\triangleq\left(\mu_{\mathrm{ADF}}-\mu_{\mathrm{PF}}\right)/\sigma_{\mathrm{PF}}$,
and the error in the posterior standard deviation estimate $\epsilon_{\sigma}\triangleq\sigma_{\mathrm{ADF}}/\sigma_{\mathrm{PF}}$,
where $\mu_{\mathrm{ADF}},\mu_{\mathrm{PF}},\sigma_{\mathrm{ADF}},\sigma_{\mathrm{PF}}$
are, respectively, the posterior mean obtained from ADF and PF, and
the posterior variance obtained from ADF and PF. The observed mean
and standard deviation of the estimation errors are $0.0018\pm0.0989$
for $\epsilon_{\mu}$ and $1.010\pm0.101$ for $\epsilon_{\sigma}$.
The results suggest that the largest errors occur in rare cases where
the posterior diverges significantly from Gaussian, and involve an
overestimation of the posterior variance. Similar results were obtained
with different parameters.

\subsection{Information gained between spikes}

The filters derived above for various population distributions differ
only in the continuous update terms, which modify the posterior between
spikes beyond the prior terms derived from the state dynamics. We
may therefore interpret this term as corresponding to information
gained by the absence of spikes. The continuous update term vanishes
in the uniform population filter of \cite{RhoSny1977} (see (\ref{eq:mean-var-uniform})),
so that in the uniform case, between spikes the posterior dynamics
are identical to the prior dynamics. This reflects the fact that lack
of spikes in a time interval is an indication that the total firing
rate is low; in the uniform population case, this is not informative,
since the total firing rate is independent of the state.

\begin{figure*}
\includegraphics[bb=0bp 0bp 534bp 390bp,width=0.4\textwidth]{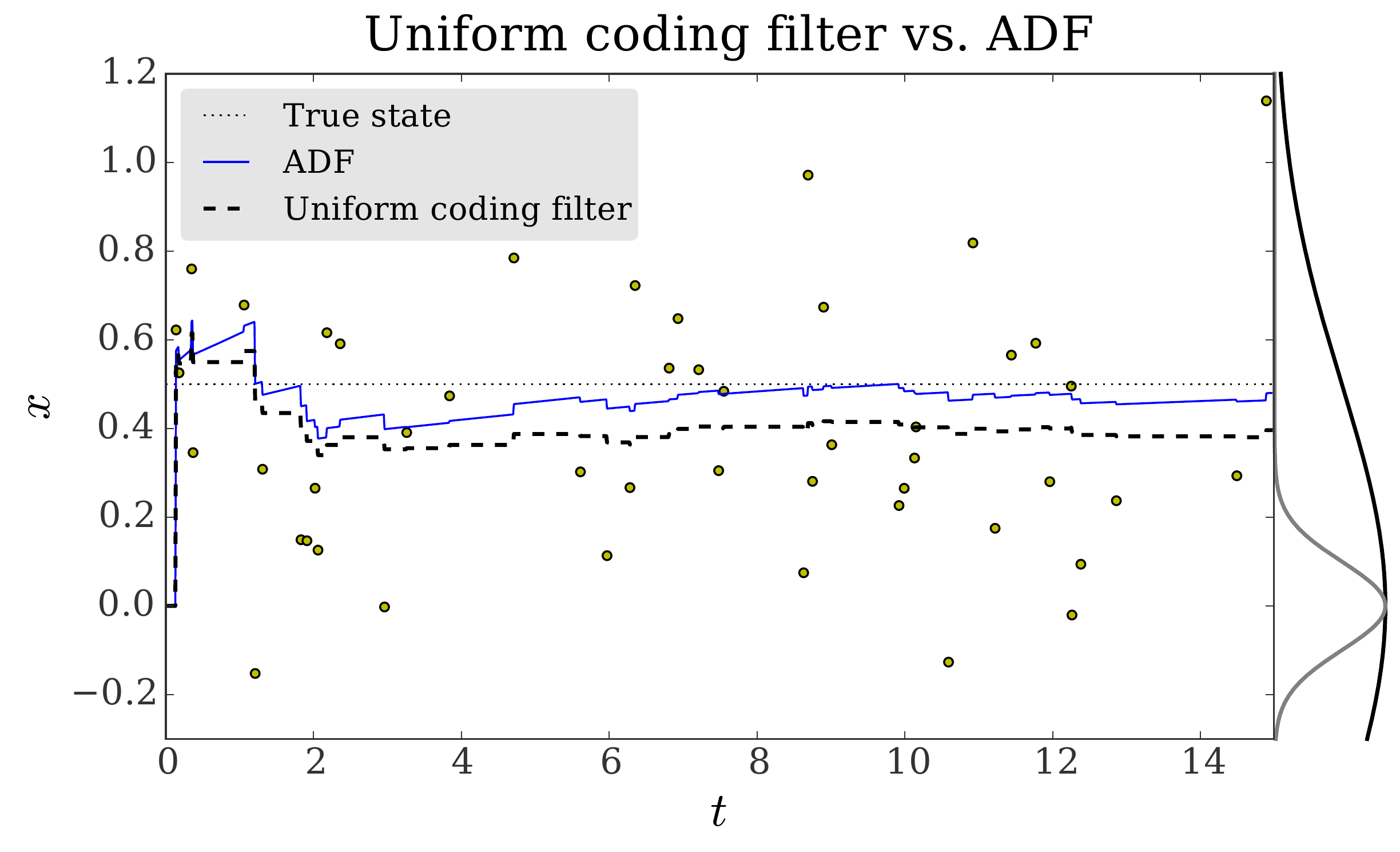}\hfill{}\includegraphics[bb=0bp 0bp 534bp 390bp,width=0.4\textwidth]{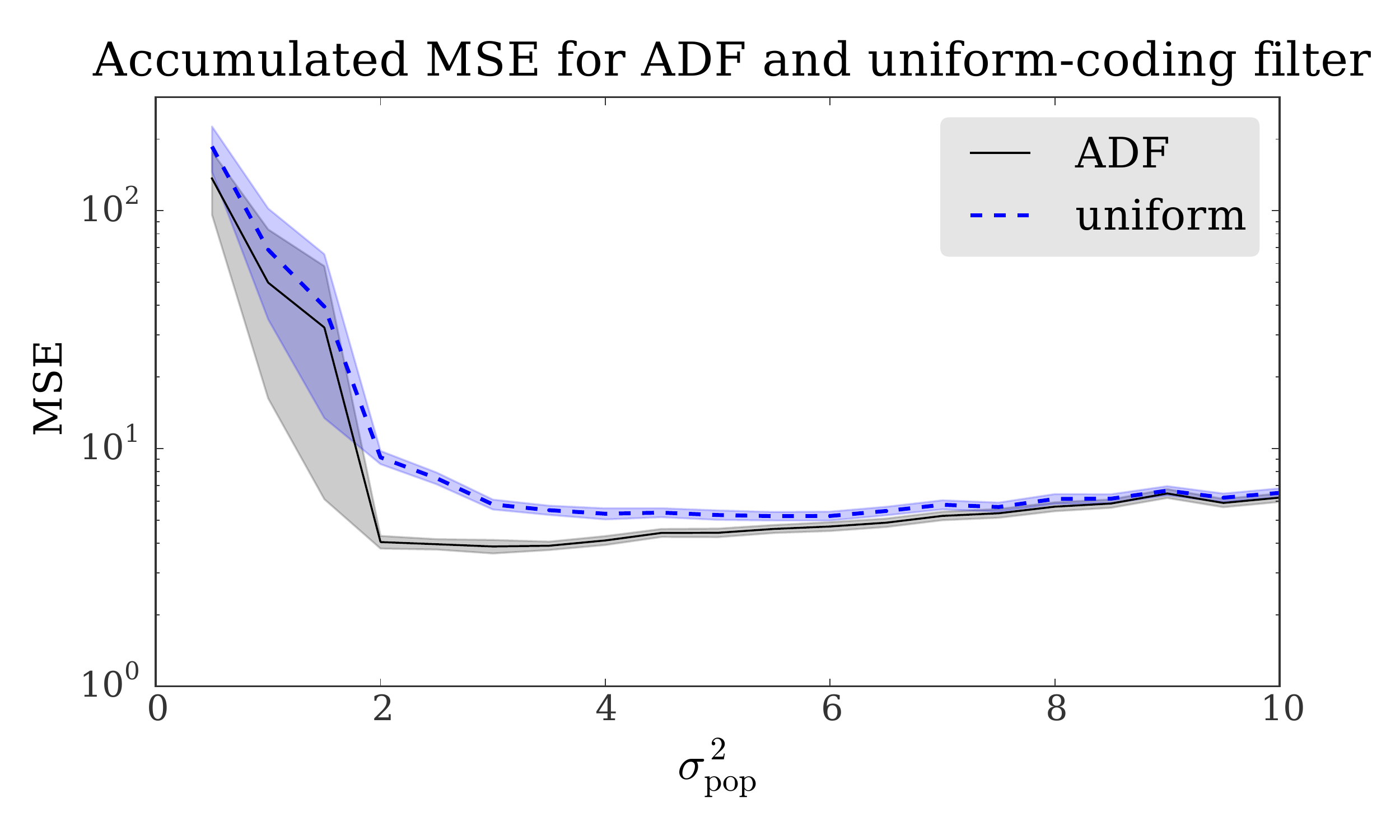}\caption{\textbf{Left} Illustration of information gained between spikes. A
static state $X_{t}=0.5$, shown in a dotted line, is observed by
a Gaussian population (\ref{eq:gaussian-f}) of Gaussian sensors (\ref{eq:guassian-sensor}),
and filtered twice: with the correct value $\sigma_{\mathrm{pop}}^{2}=0.5$
(``ADF'', solid blue line), and with $\sigma_{\mathrm{pop}}^{2}=\infty$
(``Uniform coding filter'', dashed line), which is equivalent to
ignoring the continuous update terms. Both filters are applied to
the same random realization of the observation process. The curves
to the right of each graph show the preferred stimulus density (black),
and a tuning curve centered at $\theta=0$ (gray). Both filters are
initialized with $\mu_{0}=0,\sigma_{0}^{2}=1$. \textbf{Right} Comparison
of MSE for the ADF filter and the uniform coding filter. The vertical
axis shows the integral of the square error integrated over the time
interval $\left[5,10\right]$, averaged over 1000 trials. Shaded areas
indicate estimated errors, computed as the sample standard deviation
divided by the square root of the number of trials. Parameters in
both plots are $a=d=0,c=0,\sigma_{\mathrm{pop}}^{2}=0.5,\sigma_{\mathrm{r}}^{2}=0.1,H=1,h=10$.}
\label{adf-vs-uniform}
\end{figure*}

Figure \ref{adf-vs-uniform} (left) illustrates the contribution of
the continuous update terms (\ref{eq:mean-c-gauss-gauss})-(\ref{eq:var-c-gauss-gauss})
to filter performance. A static scalar state is observed by a Gaussian
population (\ref{eq:guassian-sensor})-(\ref{eq:gaussian-f}), and
filtered twice: once with the correct value of $\sigma_{\mathrm{pop}}^{2}=\Sigma_{\mathrm{pop}}$,
and once with $\sigma_{\mathrm{pop}}\rightarrow\infty$, which yields
$\mu_{t}^{\mathrm{c}}=0$, recovering the uniform population filter
of \cite{RhoSny1977}. Between spikes, the ADF estimate moves away
from the population center $c=0$, whereas the uniform coding estimate
remains fixed. The size of this effect decreases with time, as the
posterior variance estimate (not shown) decreases. The reduction in
filtering errors gained from the continuous update terms is illustrated
in Figure \ref{adf-vs-uniform} (right). Despite the approximation
involved, the full filter significantly outperforms the uniform population
filter. The difference disappears as $\sigma_{\mathrm{pop}}$ increases
and the population becomes uniform.

\section{Encoding}

\subsection{Motivation}

We demonstrate the use of the Assumed Density Filter in determining
optimal encoding strategies, i.e., selecting the optimal sensor population
parameters $\phi$ (see Section \ref{sub:Encoding-and-decoding} above).
The study of optimal encoding has both biological and engineering
motivations. In the context of neuroscience, the optimal encoding
strategy may serve as a model for the observed characteristics of
sensory neurons, or for observed sensory adaptation (e.g., \cite{DayAbb05}).
Several works have studied optimal neural encoding using Fisher information
as a performance criterion. Fisher information is easily computed
from tuning functions in the case of static state (e.g., \cite{DayAbb05}),
and this approach completely circumvents the need to model the decoding
process. For example, in \cite{GanSim14}, the authors consider a
scalar static state observed by a parameterized population of neurons,
and analytically optimize the parameters based on Fisher information
under a constraint on the total expected rate of spikes. A few works
(e.g. \cite{YaeMei10,Susemihl2014}) consider the more difficult problem
of direct minimization of the MSE rather than the Fisher information,
by solving the corresponding filtering problem and measuring the MSE
of different encoding parameters using Monte Carlo (MC) simulations.
The filtering problem is made tractable in \cite{YaeMei10} and \cite{Susemihl2014}
by assuming a uniform population (equivalent to case 2 in Section
\ref{sub:special-cases}) above. However, sensory populations are
often non-uniform (e.g. \cite{Brand2002}), and sensory adaptation
often modifies tuning curves non-uniformly (e.g. \cite{Benucci2013}),
which motivates our more general setting. Optimization of sensor configuration
and coding is also increasingly studied in engineering contexts (e.g.,
\cite{Yuksel2013,andrievsky2010,Mourikis2006,Ny2011}), as networked
filtering and control are becoming ubiquitous. The latter studies
are usually concerned with continuous observations (often linear),
rather than PP based observations using heterogeneous biologically
motivated tuning functions as is done here.

\subsection{Encoding example}

To illustrate the use of ADF for the encoding problem, we consider
a simple example using a Gaussian population (\ref{eq:gaussian-f}).
We will study optimal encoding issues in more detail in a sequel paper. 

Previous work using a finite neuron population and a Fisher information-based
criterion \cite{HarMcAlp04} has suggested that the optimal distribution
of preferred stimuli depends on the prior variance. When it is small
relative to the tuning curve width, optimal encoding is achieved by
placing all preferred stimuli at a fixed distance from the prior mean.
On the other hand, when the prior variance is large relative to the
tuning curve width, optimal encoding is uniform (see figure 2 in \cite{HarMcAlp04}).
These results are consistent with biological observations reported
in \cite{Brand2002} concerning the encoding of aural stimuli.

\begin{figure*}
\includegraphics[bb=0bp 0bp 534bp 390bp,width=0.55\textwidth]{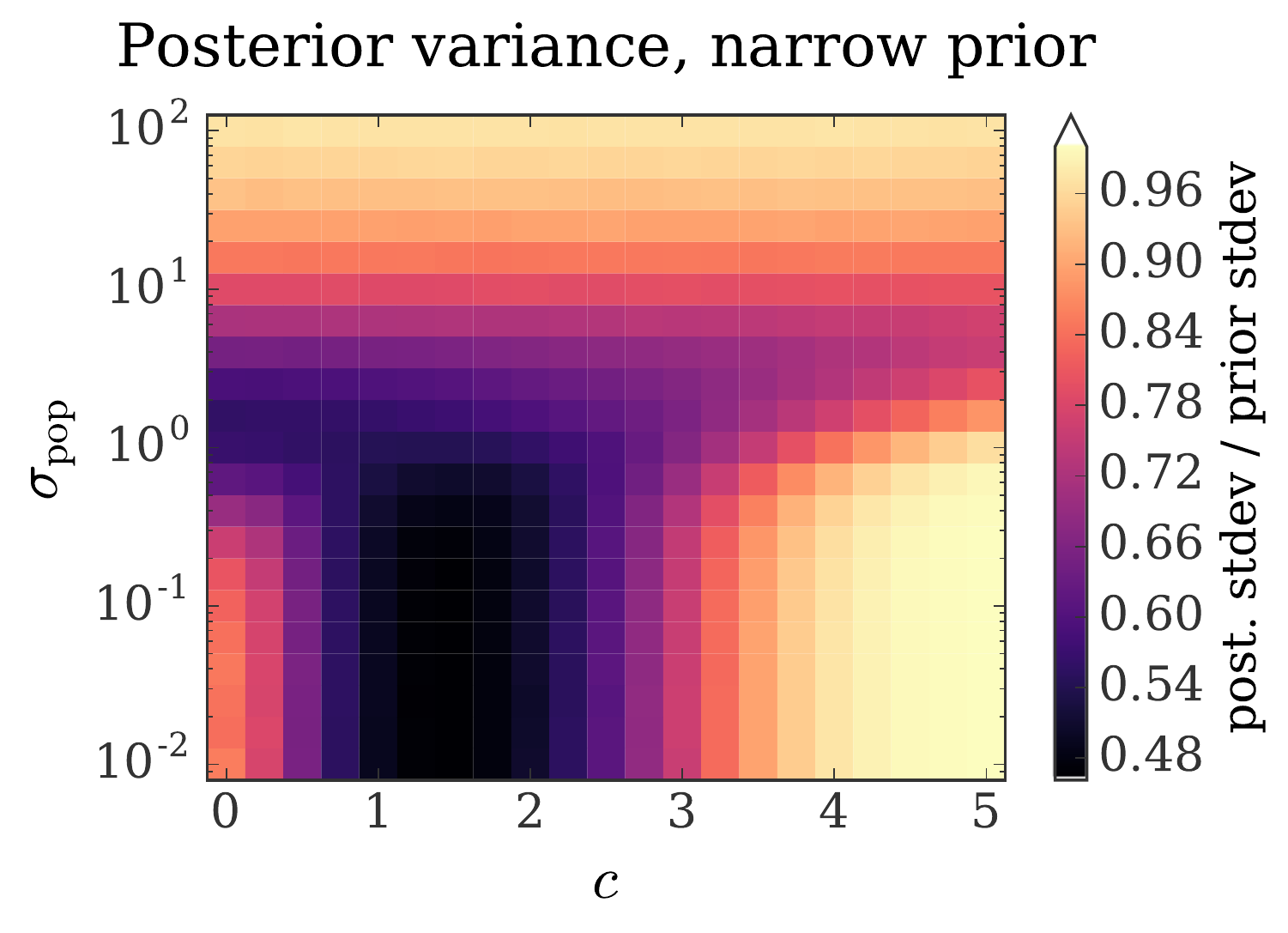}\hfill{}\includegraphics[bb=0bp 0bp 534bp 390bp,width=0.55\textwidth]{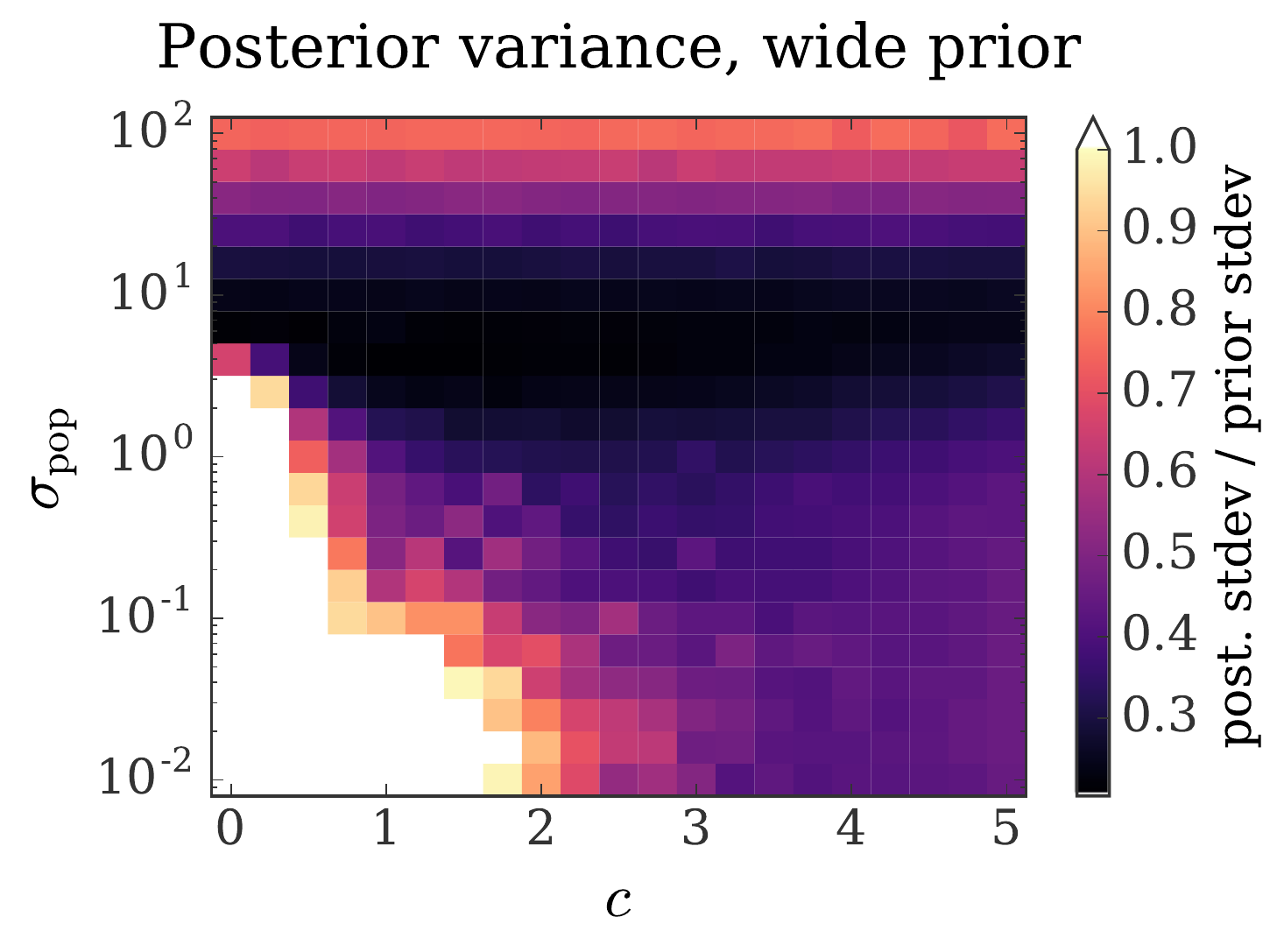}\caption{Optimal population distribution depends on prior variance relative
to tuning curve width. A scalar state with dynamics $dX_{t}=aX_{t}+d\,dW_{t}$
($a=-0.05$) is filtered with tuning curves parameters $h=50,\sigma_{\mathrm{r}}=1$
and preferred stimulus density $\mathcal{N}(c,\sigma_{\mathrm{pop}}^{2})$.
The process is initialized from its steady state distribution, $\protect\mc N\left(0,d^{2}/\left(-2a\right)\right)$.
Both graphs show the posterior standard deviation derived from the
ADF approximation, relative to the prior standard deviation $\sigma_{\mathrm{p}}$.
In the left graph, $d=0.1$ so that the prior variance is $0.1$,
whereas on the right, $d=1$, so that the prior variance is $10$.
In both cases the filter is initialized with the correct prior, and
the posterior variance is averaged over the time interval $\left[1,2\right]$
and across 1000 trials for each data point. Only non-negative values
of $c$ were simulated, but note that the process is symmetric about
zero, so that the full plots would also be symmetric about $c=0$.
The areas colored in white in the right plot correspond to parameters
where the computed posterior variance exceeded the prior variance.
This is due to poor performance of the ADF approximation for these
parameter values, in cases where no spikes occur and the true posterior
becomes bimodal.}
\label{harper}
\end{figure*}

Similar results are obtained with our model, as shown in Figure \ref{harper}.
Whereas \cite{HarMcAlp04} implicitly assumed a static state in the
computation of Fisher information, we use a time-varying scalar state.
The state obeys the dynamics
\[
dX_{t}=aX_{t}dt+d\,dW_{t}\quad\left(a<0\right),
\]
and is observed through a Gaussian population (\ref{eq:gaussian-f})
and filtered using the ADF approximation. In this case, optimal encoding
is interpreted as the simultaneous optimization of the population
center $c$ and the population variance $\Sigma_{\mathrm{pop}}$.
The process is initialized so that it has a constant prior distribution,
its variance given by $d^{2}/\left(2\left|a\right|\right)$. In Figure
\ref{harper} (left), the steady-state prior distribution is narrow
relative to the tuning curve width, leading to an optimal population
with a narrow population distribution far from the origin. In Figure
\ref{harper} (right), the prior is wide relative to the tuning curve
width, leading to an optimal population with variance that roughly
matches the prior variance.

Our approach, though more computationally expensive, offers two advantages
over the Fisher information-based method which is used in \cite{HarMcAlp04}
and which is prevalent in computational neuroscience. First, the simple
computation of Fisher information from tuning curves, commonly used
in the neuroscience literature, is based on the assumption of a static
state, whereas our method can be applied in a fully dynamic context,
including the presence of observation-dependent feedback. Second,
our approach allows the minimization of arbitrary criteria, including
the direct minimization of posterior variance or Mean Square Error
(MSE). Although, under appropriate conditions, Fisher information
approaches the MSE in the limit of infinite decoding time, it may
be a poor proxy for the MSE for finite decoding times (e.g., \cite{Bethge2002,YaeMei10}),
which are of particular importance in natural settings and in control
problems.

\section{Conclusions}

We have introduced an analytically tractable approximation to point
process filtering, allowing us to gain insight into the generally
intractable infinite-dimensional filtering problem. The approach enables
the derivation of near-optimal encoding schemes going beyond the previously
studied case of uniform population. The framework is presented in
continuous time, circumventing temporal discretization errors and
numerical imprecision in sampling-based methods, applies to fully
dynamic setups, and directly estimates the MSE rather than lower bounds
to it. It successfully explains observed experimental results, and
opens the door to many future predictions. Moreover, the proposed
strategy may lead to practically useful decoding of spike trains. 

\bibliographystyle{IEEEtran}
\bibliography{ieee_adf}

\begin{thebibliography}{10}
\providecommand{\url}[1]{#1}
\csname url@samestyle\endcsname
\providecommand{\newblock}{\relax}
\providecommand{\bibinfo}[2]{#2}
\providecommand{\BIBentrySTDinterwordspacing}{\spaceskip=0pt\relax}
\providecommand{\BIBentryALTinterwordstretchfactor}{4}
\providecommand{\BIBentryALTinterwordspacing}{\spaceskip=\fontdimen2\font plus
\BIBentryALTinterwordstretchfactor\fontdimen3\font minus
  \fontdimen4\font\relax}
\providecommand{\BIBforeignlanguage}[2]{{%
\expandafter\ifx\csname l@#1\endcsname\relax
\typeout{** WARNING: IEEEtran.bst: No hyphenation pattern has been}%
\typeout{** loaded for the language `#1'. Using the pattern for}%
\typeout{** the default language instead.}%
\else
\language=\csname l@#1\endcsname
\fi
#2}}
\providecommand{\BIBdecl}{\relax}
\BIBdecl

\bibitem{AndMoo05}
B.~Anderson and J.~Moore, \emph{Optimal Filtering}.\hskip 1em plus 0.5em minus
  0.4em\relax Dover, 2005.

\bibitem{KalBuc61}
R.~Kalman and R.~Bucy, ``New results in linear filtering and prediction
  theory,'' \emph{J. of Basic Eng., Trans. ASME, Series D}, vol. 83(1), pp.
  95--108, 1961.

\bibitem{Kalman60}
R.~Kalman, ``A new approach to linear filtering and prediction problems,''
  \emph{J. Basic Eng., Trans. ASME, Series D.}, vol. 82(1), pp. 35--45, 1960.

\bibitem{Daum05}
F.~Daum, ``Nonlinear filters: beyond the kalman filter,'' \emph{Aerospace and
  Electronic Systems Magazine, IEEE}, vol.~20, no.~8, pp. 57--69, 2005.

\bibitem{JulUhl00}
S.~Julier, J.~Uhlmann, and H.~Durrant-Whyte, ``A new method for the nonlinear
  transformation of means and covariances in filters and estimators,''
  \emph{IEEE Trans. Autom. Control}, vol. 45(3), pp. 477--482, 2000.

\bibitem{Arasaratnam2009}
I.~Arasaratnam and S.~Haykin, ``Cubature kalman filters,'' \emph{IEEE
  Transactions on automatic control}, vol.~54, no.~6, pp. 1254--1269, 2009.

\bibitem{DouJoh09}
A.~Doucet and A.~Johansen, ``A tutorial on particle filtering and smoothing:
  fifteen years later,'' in \emph{Handbook of Nonlinear Filtering}, D.~Crisan
  and B.~Rozovskii, Eds.\hskip 1em plus 0.5em minus 0.4em\relax Oxford, UK:
  Oxford University Press, 2009, pp. 656--704.

\bibitem{DayAbb05}
P.~Dayan and L.~Abbott, \emph{Theoretical Neuroscience: Computational and
  Mathematical Modeling of Neural Systems}.\hskip 1em plus 0.5em minus
  0.4em\relax MIT Press, 2005.

\bibitem{Benucci2009}
A.~Benucci, D.~L. Ringach, and M.~Carandini, ``Coding of stimulus sequences by
  population responses in visual cortex.'' \emph{Nature neuroscience}, vol.~12,
  no.~10, pp. 1317--1324, 2009.

\bibitem{GutZheKnu02}
Y.~Gutfreund, W.~Zheng, and E.~Knudsen, ``Gated visual input to the central
  auditory system.'' \emph{Science}, vol. 297, no. 5586, pp. 1556--9, Aug 2002,
  449 (Tech).

\bibitem{Benucci2013}
A.~Benucci, A.~B. Saleem, and M.~Carandini, ``Adaptation maintains population
  homeostasis in primary visual cortex.'' \emph{Nature neuroscience}, vol.~16,
  no.~6, pp. 724--9, 2013.

\bibitem{HarMcAlp04}
N.~Harper and D.~McAlpine, ``Optimal neural population coding of an auditory
  spatial cue.'' \emph{Nature}, vol. 430, no. 7000, pp. 682--686, Aug 2004,
  n1397b.

\bibitem{BobMeiEld09}
O.~Bobrowski, R.~Meir, and Y.~Eldar, ``Bayesian filtering in spiking neural
  networks: noise, adaptation, and multisensory integration.'' \emph{Neural
  Comput}, vol.~21, no.~5, pp. 1277--1320, May 2009.

\bibitem{SusMeiOpp11}
A.~Susemihl, R.~Meir, and M.~Opper, ``Analytical results for the error in
  filtering of gaussian processes,'' in \emph{Advances in Neural Information
  Processing Systems 24}, J.~Shawe-Taylor, R.~Zemel, P.~Bartlett, F.~Pereira,
  and K.~Weinberger, Eds., 2011, pp. 2303--2311.

\bibitem{SusMeiOpp13}
------, ``Dynamic state estimation based on poisson spike trains---towards a
  theory of optimal encoding,'' \emph{Journal of Statistical Mechanics: Theory
  and Experiment}, vol. 2013, no.~03, p. P03009, 2013.

\bibitem{Maybeck79}
P.~Maybeck, \emph{Stochastic Models, Estimation, and Control}.\hskip 1em plus
  0.5em minus 0.4em\relax Academic Press, 1979.

\bibitem{BriHanLeg98}
D.~Brigo, B.~Hanzon, and F.~LeGland, ``A differential geometric approach to
  nonlinear filtering: the projection filter,'' \emph{Automatic Control, IEEE
  Transactions on}, vol.~43, pp. 247--252, 1998.

\bibitem{Opper98}
M.~Opper, ``A {B}ayesian approach to online learning,'' in \emph{Online
  Learning in Neural Networks}, D.~Saad, Ed.\hskip 1em plus 0.5em minus
  0.4em\relax Cambridge university press, 1998, pp. 363--378.

\bibitem{Minka01}
T.~Minka, ``Expectation propagation for approximate bayesian inference,'' in
  \emph{Proceedings of the Seventeenth conference on Uncertainty in artificial
  intelligence}.\hskip 1em plus 0.5em minus 0.4em\relax Morgan Kaufmann
  Publishers Inc., 2001, pp. 362--369.

\bibitem{SusemihlThesis2014}
A.~Susemihl, ``Optimal population coding of dynamic stimuli,'' Ph.D.
  dissertation, Technical University Berlin, 2014.

\bibitem{Snyder1972}
D.~Snyder, ``{Filtering and detection for doubly stochastic Poisson
  processes},'' \emph{IEEE Transactions on Information Theory}, vol.~18, no.~1,
  pp. 91--102, Jan. 1972.

\bibitem{Segall1976}
A.~Segall, ``Recursive estimation from discrete-time point processes,''
  \emph{Information Theory, IEEE Transactions on}, vol.~22, no.~4, pp.
  422--431, 1976.

\bibitem{RhoSny1977}
I.~Rhodes and D.~Snyder, ``{Estimation and control performance for space-time
  point-process observations},'' \emph{IEEE Transactions on Automatic Control},
  vol.~22, no.~3, pp. 338--346, 1977.

\bibitem{Snyder1977}
D.~Snyder, I.~Rhodes, and E.~Hoversten, ``A separation theorem for stochastic
  control problems with point-process observations,'' \emph{Automatica},
  vol.~13, no.~1, pp. 85--87, 1977.

\bibitem{Segall1978}
A.~Segall, ``Centralized and decentralized control schemes for
  {Gauss}-{Poisson} processes,'' \emph{IEEE Transactions on Automatic Control},
  vol.~23, no.~1, pp. 47--57, 1978.

\bibitem{Bremaud81}
P.~Br\'emaud, \emph{Point Processes and Queues: Martingale Dynamics}.\hskip 1em
  plus 0.5em minus 0.4em\relax Springer, New York, 1981.

\bibitem{Frey2001}
R.~Frey and W.~J. Runggaldier, ``A nonlinear filtering approach to volatility
  estimation with a view towards high frequency data,'' \emph{International
  Journal of Theoretical and Applied Finance}, vol.~4, no.~02, pp. 199--210,
  2001.

\bibitem{Harel2015}
Y.~Harel, R.~Meir, and M.~Opper, ``A tractable approximation to optimal point
  process filtering: Application to neural encoding,'' in \emph{Advances in
  Neural Information Processing Systems}, 2015, pp. 1603--1611.

\bibitem{YaeMei10}
S.~Yaeli and R.~Meir, ``Error-based analysis of optimal tuning functions
  explains phenomena observed in sensory neurons.'' \emph{Front Comput
  Neurosci}, vol.~4, p. 130, 2010.

\bibitem{Susemihl2014}
A.~Susemihl, R.~Meir, and M.~Opper, ``{Optimal Neural Codes for Control and
  Estimation},'' \emph{Advances in Neural Information Processing Systems}, pp.
  1--9, 2014.

\bibitem{Segall1975-modelling}
A.~Segall and T.~Kailath, ``The modeling of randomly modulated jump
  processes,'' \emph{Information Theory, IEEE Transactions on}, vol.~21, no.~2,
  pp. 135--143, 1975.

\bibitem{Oksendal2003}
B.~{\O}ksendal, \emph{Stochastic {Differential} {Equations}}.\hskip 1em plus
  0.5em minus 0.4em\relax Springer, 2003.

\bibitem{GanSim14}
D.~Ganguli and E.~Simoncelli, ``Efficient sensory encoding and bayesian
  inference with heterogeneous neural populations.'' \emph{Neural Comput},
  vol.~26, no.~10, pp. 2103--2134, 2014.

\bibitem{Brand2002}
A.~Brand, O.~Behrend, T.~Marquardt, D.~McAlpine, and B.~Grothe, ``{Precise
  inhibition is essential for microsecond interaural time difference coding.}''
  \emph{Nature}, vol. 417, no. 6888, pp. 543--547, 2002.

\bibitem{Yuksel2013}
S.~Y{\"u}ksel and T.~Ba{\c{s}}ar, \emph{Stochastic networked control systems:
  Stabilization and optimization under information constraints}.\hskip 1em plus
  0.5em minus 0.4em\relax Springer Science \& Business Media, 2013.

\bibitem{andrievsky2010}
B.~R. Andrievsky, A.~S. Matveev, and A.~L. Fradkov, ``Control and estimation
  under information constraints: Toward a unified theory of control,
  computation and communications,'' \emph{Automation and Remote Control},
  vol.~71, no.~4, pp. 572--633, 2010.

\bibitem{Mourikis2006}
A.~I. Mourikis and S.~I. Roumeliotis, ``Optimal sensor scheduling for
  resource-constrained localization of mobile robot formations,'' \emph{IEEE
  Transactions on Robotics}, vol.~22, no.~5, pp. 917--931, Oct 2006.

\bibitem{Ny2011}
J.~L. Ny, E.~Feron, and M.~A. Dahleh, ``Scheduling continuous-time kalman
  filters,'' \emph{IEEE Transactions on Automatic Control}, vol.~56, no.~6, pp.
  1381--1394, June 2011.

\bibitem{Bethge2002}
M.~Bethge, D.~Rotermund, and K.~Pawelzik, ``Optimal short-term population
  coding: when fisher information fails.'' \emph{Neural Comput}, vol.~14,
  no.~10, pp. 2317--2351, Oct 2002.

\end{thebibliography}

\end{document}